\documentclass[10pt,journal,compsoc]{IEEEtran}
\ifCLASSOPTIONcompsoc
  \usepackage[nocompress]{cite}
\else
  \usepackage{cite}
\fi
\usepackage{epsfig}
\usepackage{graphicx}
\usepackage{amsmath}
\usepackage{amssymb}
\usepackage{flexisym}
\usepackage{breqn}
\usepackage{subfigure}
\usepackage{soul}
\usepackage[linesnumbered,ruled]{algorithm2e}
\usepackage[table]{xcolor}
\usepackage{colortbl}
\definecolor{Gray}{gray}{0.85}
\usepackage{array,multirow,graphicx}
\usepackage{xfrac}
\usepackage{booktabs}

\newcommand{\tr}{{\rm tr}}
\newcommand{\range}{{\sl range}}
\newcommand{\lspan}{{\sl span}}

\begin{document}

\title{A Novel Geometric Framework on Gram Matrix Trajectories for Human Behavior Understanding}

\author{Anis~Kacem, Mohamed~Daoudi, Boulbaba~Ben~Amor, Stefano~Berretti, and~Juan~Carlos~Alvarez-Paiva
\IEEEcompsocitemizethanks{\IEEEcompsocthanksitem A. Kacem, M. Daoudi, and B. Ben~Amor are with IMT Lille Douai and CRIStAL laboratory, CNRS-UMR-9189, France. \protect
\IEEEcompsocthanksitem S. Berretti is with the University of Florence, Italy.
\IEEEcompsocthanksitem J.C. Alvarez-Paiva is with the University of Lille and Painlev\'e laboratory, CNRS-UMR-8524, France.}}

\IEEEtitleabstractindextext{%
\begin{abstract}
In this paper, we propose a novel space-time geometric representation of human landmark configurations and derive tools for comparison and classification. We model the temporal evolution of landmarks as parametrized trajectories on the Riemannian manifold of positive semidefinite matrices of fixed-rank. Our representation has the benefit to bring naturally a second desirable quantity when comparing shapes -- the spatial covariance -- in addition to the conventional affine-shape representation. We derived then geometric and computational tools for rate-invariant analysis and adaptive re-sampling of trajectories, grounding on the Riemannian geometry of the underlying manifold. Specifically, our approach involves three steps: (1) landmarks are first mapped into the Riemannian manifold of positive semidefinite matrices of fixed-rank to build time-parameterized trajectories; (2) a temporal warping is performed on the trajectories, providing a geometry-aware (dis-)similarity measure between them; (3) finally, a pairwise proximity function SVM is used to classify them, incorporating the (dis-)similarity measure into the kernel function. We show that such representation and metric achieve competitive results in applications as action recognition and emotion recognition from 3D skeletal data, and facial expression recognition from videos. Experiments have been conducted on several publicly available up-to-date benchmarks.

\end{abstract}

}

\maketitle

\IEEEdisplaynontitleabstractindextext

\IEEEpeerreviewmaketitle

\IEEEraisesectionheading{\section{Introduction}\label{sect:introduction}}
\IEEEPARstart{S}{everal} human-related Computer Vision problems can be approached by first detecting and tracking landmarks from visual data. A relevant example of this is given by the estimated 3D location of the joints of the skeleton in depth streams~\cite{shotton2013real}, and their use in action and daily activity recognition. More sophisticated solutions for automatic tracking of the skeleton do exist, as the IR body markers used in MoCap systems, but they are expensive in cost and time.
Another relevant example is represented by the face, for which several approaches have been proposed for fiducial points detection and tracking in video~\cite{AsthanaZCP14,XiongT13}.
These techniques generate temporal sequences of landmark configurations, which exhibit several variations due to affine and projective transformations, inaccurate detection and tracking, missing data, etc.
While there have been many efforts in the analysis of temporal sequences of landmarks, the problem is far from being solved and the current solutions are facing many technical and practical problems. For instance, many general techniques for temporal sequence analysis rely on computing the Euclidean distance between two temporal sequences and do not take into account the implicit dynamics of the sequences~\cite{Zafeirioucvpr2016}. %
In practice, when analyzing the temporal dynamics of landmark configurations, there are four main aspects to deal with that require us to define:
(1) A shape representation invariant to undesirable transformations; (2) A temporal modeling of landmark sequences; (3) A suitable rate-invariant distance between arbitrary sequences, and (4) A solution for temporal sequence classification.

\par
\begin{figure}[!t]
\centering
\includegraphics[width =\linewidth]{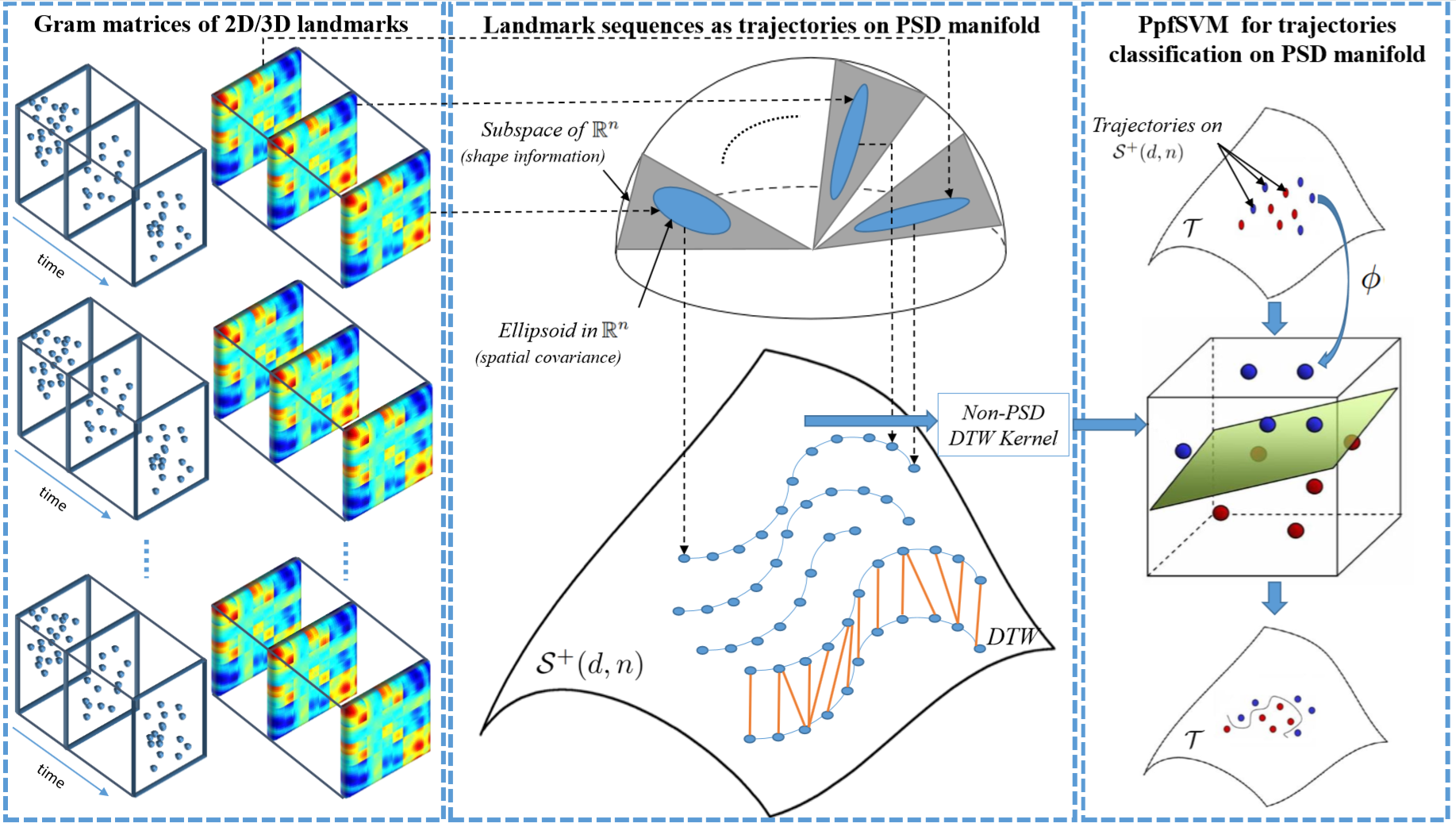}
\caption{Overview of the proposed approach. Given a landmark sequence, the Gram matrices are computed for each landmark configuration to build trajectories on $\mathcal{S}^+(d,n)$. A moving shape is hence assimilated to an ellipsoid traveling along $d$-dimensional subspaces of $\mathbb{R}^n$, with $d_{\mathcal{S}^+}$ used to compare static ellipsoids. Dynamic Time Warping (DTW) is then used to align and compare trajectories in a rate-invariant manner. Finally, the ppfSVM is used on these trajectories for classification.
\label{Fig:AppOverview}
}
\end{figure}

 In this paper, we propose a method that effectively model the comparison and classification of temporal sequences of landmarks. In doing so, we define new solutions for the four points listed above. Considering the first issue, we propose a novel shape modeling, invariant to rigid motion, by embedding shapes represented with their corresponding Gram matrix into the Positive Semidefinite Riemannian manifold. Such representation has the advantage of bringing naturally a second desirable quantity when comparing shapes -- the spatial covariance -- in addition to the conventional affine-shape representation.
For the second issue, that is to model the dynamics and dependency relations in both temporal and spatial domains, we represent the temporal evolution of landmarks as parametrized trajectories on the Riemannian manifold of positive semidefinite matrices of fixed-rank. For what concerns the third issue, geometric and computational tools for rate-invariant analysis and adaptive re-sampling of trajectories, grounding on the Riemannian geometry of the manifold, are proposed. Finally, a variant of SVM that takes into account the nonlinearity of this space is proposed for trajectory classification.
An overview of the full approach is given in Fig.~\ref{Fig:AppOverview}.

A preliminary version of this work appeared in~\cite{KacemICCV2017}, with application to facial expression recognition using $2$D landmarks. In this work, we generalize the idea significantly, by considering new applications using $3$D landmarks. In summary, the main contributions of this work are:
\begin{itemize}
\item A novel static shape representation based on Gramian matrices of centered $2$D and $3$D landmark configurations. A comprehensive study of the Riemannian geometry of the space of representations, termed the cone of Positive Semi-Definite $n \times n$ matrices of fixed-rank $d$ is conducted ($d=2$ or $d=3$ for $2$D or $3$D landmark configurations, respectively). Despite the large use of these matrices in several research fields, to our knowledge, this is the first application in static and dynamic shape analysis.

\item In addition to the affine-invariant analysis, our representation brings a spatial covariance of the landmarks. In comparison to the preliminary version of this work appeared in~\cite{KacemICCV2017}, the proposed framework has been extended to study trajectories of 3D landmarks. The effectiveness of the Gramian representation, \emph{i.e.}, including the spatial covariance, compared to the Grassmannian was confirmed by experiments of 3D action and emotion recognition.

\item A temporal extension of the representation via parametrized trajectories in the underlying Riemannian manifold, with associated computational tools for temporal alignment and adaptive re-sampling of trajectories.

\item A solution for trajectory classification based on pairwise proximity function SVM (ppfSVM), where pairwise (dis-)similarity measures between trajectories are computed using the metric of the underlying manifold. 

\item Extensive experiments of our framework in three applications -- $3$D action recognition, emotion recognition from $3$D human motion, and $2$D facial expression recognition -- demonstrate its competitiveness with respect to the state-of-the-art.
\end{itemize}

The remaining of the paper is organized as follows:
In Section~\ref{sect:related-work}, we discuss on the related works that use trajectories to model the temporal dynamics in different application contexts; In Section~\ref{sect:representation}, we propose a method to represent static landmark configurations using the Gram matrix, and also provide a comprehensive study of the Riemannian geometry of the space of these matrices; Based on this mathematical background, in Section~\ref{sect:temporal-modeling}, temporal sequences of landmarks are modeled as trajectories on the Gramian manifold, and a suitable measure for comparing trajectories on the manifold is proposed. The classification of the trajectories on the manifold is described in Section~\ref{sect:classification}. An extensive experimental validation of the proposed approach is reported in Section~\ref{sect:results}. Experiments account for different application contexts, including $3$D human action recognition, emotion recognition from $3$D body movement and $2$D facial expression recognition, also comparing with state-of-the-art solutions. Finally, conclusions are drawn in Section~\ref{sect:conclusion}.

\section{Related Work}
\label{sect:related-work}
In the following, we first review essential literature solutions proposed for modeling the temporal evolution of landmark sequences, then we focus on mapping static shape representations as trajectories on Riemannian manifolds. Note that, often, shape-preserving transformations are filtered out from the static representation, and the rate-invariance is proposed at the trajectory level.

\medskip

\par \noindent \textbf{A. Temporal Modeling of Landmark Sequences} -- In the work of Slama~\emph{et al.}~\cite{SlamaPattern15}, a temporal sequence was represented as a Linear Dynamical System (LDS). The observability matrix of the LDS was then approximated by a finite matrix~\cite{TuragaPAMI2011}. The subspace spanned by the columns of this finite observability matrix corresponds to a point on a Grassmann manifold. Thus, the LDS is represented at each time-instant as a point on the Grassmann manifold. Each video sequence is modeled as an element of the Grassmann manifold, and action learning and recognition is cast to a classification problem on this manifold. Proximity between two spatio-temporal sequences is measured by a distance between two subspaces on the Grassmann manifold.
Huang~\emph{et al.}~\cite{Huang_2016_CVPR} formulated the LDS as an infinite Grassmann manifold, and Venkataraman~\emph{et al.}~\cite{Venkataraman16} proposed a shape-theoretic framework for analysis of non-linear dynamical systems. Applications were shown to activity recognition using motion capture and RGB-D sensors, and to activity quality assessment for stroke rehabilitation. Taking a different direction, the authors of~\cite{cavazza2016kernelized} and~\cite{wang2015beyond} proposed to map full skeletal sequences onto SPD manifolds. That is, given an arbitrary sequence, it is summarized by a covariance matrix derived from the velocities computed from neighboring frames or from the 3D landmarks themselves, respectively. In both of these works kernelized versions of covariance matrices are considered. Zhang~\emph{et al.}~\cite{Zhang_2016_CVPR} represented temporal landmark sequences using regularized Gram matrices derived from the Hankel matrices of landmark sequences. The authors show that the Hankel matrix of a $3$D landmark sequence is related to an Auto-Regressive (AR) model \cite{li2012cross}, where only the linear relationships between landmark static observations are captured. The Gram matrix of the Hankel matrix is computed to reduce the noise and is seen as a point on the positive semi-definite manifold. To analyze/compare the Gram matrices, they regularized their ranks resulting in positive definite matrices and considered metrics on the positive definite manifold. 

Several solutions have experimented the application of Recurrent Neural Networks (RNNs) and Long Short Term Memory (LSTM) networks to the case of 3D landmarks represented by the joints of the human skeleton for 3D human action recognition. In fact, human actions can be interpreted as time series of body configurations, which can be represented in an effective and compact way by the 3D locations of the joints of the skeleton. In this way, each video sample can be modeled as a sequential representation of configurations. 
This approach was followed by Veeriah~\emph{et al.}~\cite{veeriah:2015} who presented a family of differential RNNs (dRNNs) that extend LSTM by a new gating mechanism to extract the derivatives of the internal state (DoS). The DoS was fed to the LSTM gates to learn salient dynamic patterns in 3D skeleton data.
Du~\emph{et al.}~\cite{du:2015} proposed an end-to-end hierarchical RNN for skeleton based action recognition. First, the human skeleton was divided into five parts, which are then feed to five subnets. As the number of layers increases, the representations in the subnets are hierarchically fused to be the inputs of higher layers. The final representations of the skeleton sequences are fed into a single-layer perceptron, and the temporally accumulated output of the perceptron is the final decision. To ensure effective learning of the deep model, Zhu~\emph{et al.}~\cite{zhu:2016}
designed an in-depth dropout algorithm for the LSTM neurons in the last layer, which helps the network to learn complex motion dynamics. To further regularize the learning, a co-occurrence inducing norm was added to the network’s cost function, which enforced the learning of groups of co-occurring and discriminative joints.
A part aware LSTM model was proposed by Shahroudy~\emph{et al.}~\cite{shahroudy:2016} to utilize the physical structure of the human body to improve the performance of the LSTM learning framework.
Instead of keeping a long-term memory of the entire body’s motion in the cell, this is split to part-based cells. In this way, the context of each body part is kept independently, and the output of the part based LSTM (P-LSTM) unit is represented as a combination of independent body part context information. Each part cell has therefore its individual input, forget, and modulation gates, but the output gate is shared among the body parts.
A broader review of methods that apply deep architectures can be found in the survey by Ioannidou~\emph{et al.}~\cite{ioannidou:2017} for the case of generic 3D data, and in the survey by Wang~\emph{et al.}~\cite{wang:2017} and Berretti~\emph{et al.}~\cite{Berretti:2018} for the particular case of human action recognition from 3D data.

\medskip

\par \noindent \textbf{B. Analyzing Shape Trajectories on Riemannian Manifolds} --
One promising idea is to formulate the motion features as trajectories. Matikainen~\emph{et al.}~\cite{Matikainen_2009_6466} presented a method for using the trajectories of tracked feature points in a bag of words paradigm for video action recognition. Despite of the promising results obtained, the authors did not take into account the geometric information of the trajectories.
More recently, in the case of human skeleton in RGB-Depth images, Devanne~\emph{et al.}~\cite{Devanne2015Cybernetics}
proposed to formulate the action recognition task as the problem of computing a distance between trajectories generated by the joints moving during the action. An action is then interpreted as a parameterized curve on the hyper-sphere of the human skeleton. However, this approach does not take into account the relationship between the joints. In the same direction, Su~\emph{et al.}~\cite{Su2014} proposed a metric that considers the time-warping on a Riemannian manifold, thus allowing trajectories registration and the computation of statistics on the trajectories. Su~\emph{et al.}~\cite{Su_2014_CVPR} applied this framework to the problem of visual speech recognition. Similar ideas have been developed by Ben Amor~\emph{et al.}~\cite{Boulbaba2016PAMI} on the Kendall's shape space with application to action recognition using rate-invariant analysis of skeletal shape trajectories.

Anirudh~\emph{et al.}~\cite{AnirudhTSS17} started from the framework of Transported Square-Root Velocity Fields (TSRVF), which has desirable properties including a rate-invariant metric and vector space representation. Based on this framework, they proposed to learn an embedding such that each action trajectory is mapped to a single point in a low-dimensional Euclidean space, and the trajectories that differ only in temporal rates map to the same point. The TSRVF representation and accompanying statistical summaries of Riemannian trajectories are used to extend existing coding methods such as PCA, KSVD, and Label Consistent KSVD to Riemannian trajectories. In the experiments, it is shown such coding efficiently captures trajectories in action recognition, stroke rehabilitation, visual speech recognition, clustering, and diverse sequence sampling.

In~\cite{vemulapalli2014human}, Vemulapalli~\emph{et al.} proposed a Lie group trajectory representation of the skeletal data on the product space of Special Euclidean ($SE$) groups. For each frame, the latter representation is obtained by computing the Euclidean transformation matrices encoding rotations and translations between different joint pairs. The temporal evolution of these matrices is seen as a trajectory on $SE(3) \times \cdots \times SE(3)$ and mapped to the tangent space of a reference point. A one-versus-all SVM, combined with Dynamic Time Warping and Fourier Temporal Pyramid (FTP) is used for classification. One limitation of this method is that mapping trajectories to a common tangent space using the logarithm map could result in significant approximation errors. Aware of this limitation, the same authors proposed in~\cite{vemulapalli2016rolling} a mapping combining the usual logarithm map with a rolling map that guarantees a better flattening of trajectories on Lie groups.

\section{Representation of Static Landmark Configurations}
\label{sect:representation}

Let us consider an arbitrary sequence of landmark configurations $\{Z_0,\ldots, Z_{\tau}\}$. Each configuration $Z_{i}$ $(0 \leq i\leq \tau)$ is an $n \times d$ matrix of rank $d$ encoding the positions of $n$ distinct landmark points in $d$ dimensions.
In our applications, we only consider the configurations of landmark points in two- or three-dimensional space (\emph{i.e.}, $d$=2 or $d$=3) given by, respectively, $p_1 = (x_1,y_1), \ldots, p_n = (x_n,y_n)$ or $p_1 = (x_1,y_1,z_1), \ldots, p_n = (x_n,y_n,z_n)$. We are interested in studying such sequences or curves of landmark configurations up to Euclidean motions. In the following, we will first propose a representation for static observations, then adopt a time-parametrized representation for temporal analysis.

As a first step, we seek a shape representation that is invariant up to Euclidean transformations (rotation and translation). Arguably, the most natural choice is the matrix of pairwise distances between the landmarks of the same shape augmented by the distances between all the landmarks and their center of mass $p_0$. Since we are dealing with Euclidean distances, it will turn out to be more convenient to consider the matrix of the squares of these distances. Also note that by subtracting the center of mass from the coordinates of the landmarks, these can be considered as \textit{centered}: the center of mass is always at the origin. From now on, we will assume $p_0=(0,0)$ for $d=2$ (or $p_0=(0,0,0)$ for $d=3$). With this provision, the augmented pairwise square-distance matrix $\mathcal{D}$ takes the form,

$$
\centering
\mathcal{D} :=
\begin{pmatrix}
0 & \|p_1\|^2 &\cdots & \|p_n\|^2 \\
\|p_1\|^2 & 0 & \cdots & \|p_1 - p_n\|^2 \\
\vdots & \vdots & \vdots & \vdots  \\
\|p_n\|^2 & \|p_n - p_1\|^2 & \cdots & 0 \\
\end{pmatrix} \; ,
$$
\noindent where $\| \cdot \|$ denotes the norm associated to the $l^2$-inner product $\langle \cdot , \cdot \rangle$.
A key observation is that the matrix $\mathcal{D}$ can be easily obtained from the $n \times n$ Gram matrix $G := ZZ^T$. Indeed, the entries of $G$ are the pairwise inner products of the points $p_1, \ldots, p_n$,
\begin{equation}
\label{eq:gram}
G=ZZ^T = \langle p_i, p_j\rangle, \; \  1 \leq i, j \leq n \; ,
\end{equation}

\noindent and the equality
\begin{equation}
\mathcal{D}_{ij} =
\langle p_i, p_i \rangle
- 2 \langle p_i, p_j \rangle
+ \langle p_j, p_j \rangle, \; \ 0 \leq i, j \leq n \; ,
\end{equation}

\noindent establishes a linear equivalence between the set of $n \times n$ Gram matrices and the augmented square-distance $(n+1) \times (n+1)$ matrices of distinct landmark points. On the other hand, Gram matrices of the form $ZZ^T$, where $Z$ is an $n \times d$ matrix of rank $d$ are characterized as $n \times n$ positive semidefinite matrices of rank $d$. For a detailed discussion of the relation between positive semidefinite matrices, Gram matrices, and square-distance matrices, we refer the reader to Section~6.2.1 of~\cite{deza2009geometry}. Conveniently for us, the Riemannian geometry of the space of these matrices, called the positive semidefinite cone $\mathcal{S}^+(d,n)$, was studied in~\cite{Bonnabel2009SIAM,faraki2016image,meyer2011regression,vandereycken2009embedded}. An alternative shape representation, considered in~\cite{BegelforW06} and~\cite{Taheri2001FG}, associates to each configuration $Z$ the $d$-dimensional subspace $\lspan(Z)$ spanned by its columns. This representation, which exploits the well-known geometry of the Grassmann manifold $\mathcal{G}(d,n)$ of $d$-dimensional subspaces in $\mathbb{R}^n$ is invariant under {\it all} invertible linear transformations. By fully encoding the set of all mutual distances between landmark points, the Euclidean shape representation proposed in this paper supplements the affine shape representation with the knowledge of the $d \times d$ covariance matrix for the centered landmarks that lie on the manifold of Symmetric Positive Definite (SPD) matrices. This leads to considerable improvements in the results of the conducted experiments.

\subsection{Riemannian Geometry of $\mathcal{S}^+(d,n)$}\label{sect:riemannian-geometry}
Given an $n \times d$ matrix $Z$ of rank $d$, its polar decomposition $Z = UR$ with $R = (Z^T Z)^{1/2}$ allows us to write the Gram matrix $ZZ^T$ as $UR^2U^T$. Since the columns of the matrix $U$ are orthonormal, this decomposition defines a map
\begin{alignat*}{2}
\Pi : & \mathcal{V}_{d,n} \times  \mathcal{P}_d \rightarrow \mathcal{S}^+(d,n) \\
  &(U,R^2)\mapsto UR^2U^T \; ,
\end{alignat*}

\noindent from the product of the Stiefel manifold $\mathcal{V}_{d,n}$ and the cone of $d \times d$ positive definite matrices $\mathcal{P}_d$ to the manifold $\mathcal{S}^+(d,n)$ of $n \times n$ positive semidefinite matrices of rank $d$. The map $\Pi$ defines a principal fiber bundle over $\mathcal{S}^+(d,n)$ with fibers
$$
\Pi^{-1}(UR^2U^T) = \{(UO, O^TR^2O) : O \in \mathcal{O}(d) \} \; ,
$$

\noindent where $\mathcal{O}(d)$ is the group of $d \times d$ orthogonal matrices.
Bonnabel and Sepulchre~\cite{Bonnabel2009SIAM} used this map and the geometry of the {\it structure space} $\mathcal{V}_{d,n} \times \mathcal{P}_d$ to introduce a Riemannian metric on $ \mathcal{S}^+(d,n)$
and study its geometry.

\subsection{Tangent Space and Riemannian Metric}\label{sect:tangent-space}

The tangent space $T_{(U,R^2)}(\mathcal{V}_{d,n} \times \mathcal{P}_d)$ consists of pairs $(M,N)$, where $M$ is a $n \times d$ matrix satisfying $M^TU + U^TM = 0$ and $N$ is any $d \times d$ symmetric matrix. Bonnabel and Sepulchre defined a {\it connection} (see~\cite[p.~63]{Kobayashi-Nomizu:1963}) on the principal bundle $\Pi : \mathcal{V}_{d,n} \times  \mathcal{P}_d \rightarrow \mathcal{S}^+(d,n)$ by setting the horizontal subspace $\mathcal{H}_{(U,R^2)}$ at the point $(U,R^2)$ to be the space of tangent vectors $(M,N)$ such that $M^TU = 0$ and $N$ is an arbitrary $d \times d$ symmetric matrix. They also defined an inner product on $\mathcal{H}_{(U,R^2)}$: given two tangent vectors $A=(M_1, N_1)$ and $B= (M_2, N_2)$ on $\mathcal{H}_{(U,R^2)}$, set
\begin{equation}
\centering
\langle (A,B) \rangle_{\mathcal{H}_{U,R^2}}=\tr(M_1^TM_2) + k \ \tr(N_1R^{-2}N_2R^{-2}) \; ,
\label{eq:Rmetrics}
\end{equation}

\noindent where $k > 0$ is a real parameter.

It is easily checked that the action of the group of $d \times d$ orthogonal matrices on the fiber $\Pi^{-1}(UR^2U^T)$ sends horizontals to horizontals isometrically. It follows that the inner product on $T_{UR^2U^T}\mathcal{S}^+(d,n)$ induced from that of $\mathcal{H}_{(U,R^2)}$ via the linear isomorphism $D\Pi$ is independent of the choice of point $(U,R^2)$ projecting onto $UR^2U^T$. This procedure defines a Riemannian metric on $\mathcal{S}^+(d,n)$ for which the natural projection
\begin{alignat*}{2}
  \rho : \ &  \mathcal{S}^+(d,n) \rightarrow \mathcal{G}(d,n) \\
  &G\mapsto \range(G) \; ,
\end{alignat*}

\noindent is a Riemannian submersion. This allows us to relate the geometry of $\mathcal{S}^+(d,n)$ with that of the Grassmannian $\mathcal{G}(d,n)$.

Recall that the geometry of the Grassmannian $\mathcal{G}(d,n)$ is easily described by using the map
$$
\lspan : V_{d,n} \rightarrow \mathcal{G}(d,n) \; ,
$$

\noindent that sends an $n \times d$ matrix with orthonormal columns $U$ to their span $\lspan(U)$. Given two subspaces $\mathcal{U}_1 = \lspan(U_1)$ and $\mathcal{U}_2=\lspan(U_2) \in \mathcal{G}(d,n)$, the geodesic curve connecting them is
\begin{dmath}
\label{eq:GeoGrass}
\lspan(U(t))= \lspan(U_{1}\cos(\Theta t)+M\sin(\Theta t)) \; ,
\end{dmath}

\noindent where $\Theta$ is a $d \times d$ diagonal matrix formed by the \textit{principal angles} between $\mathcal{U}_1$ and $\mathcal{U}_2$, while the matrix $M$ is given by the formula $M=(I_n-U_{1}U_1^T)U_2 F$, with $F$ being the pseudoinverse $diag(\sin(\theta_1),\sin(\theta_2))$. The Riemannian distance between $\mathcal{U}_1$ and $\mathcal{U}_2$ is given by
\begin{equation}
d^2_{\mathcal{G}}(\mathcal{U}_1,\mathcal{U}_2)=\|\Theta\|^2_F \;  .
\label{eq:distGrass}
\end{equation}

\subsection{Pseudo-Geodesics and Closeness in $\mathcal{S}^+(d,n)$}\label{sect:pseudo-geodesics}
\begin{figure}[h]
\centering
\includegraphics[width=.8\linewidth]{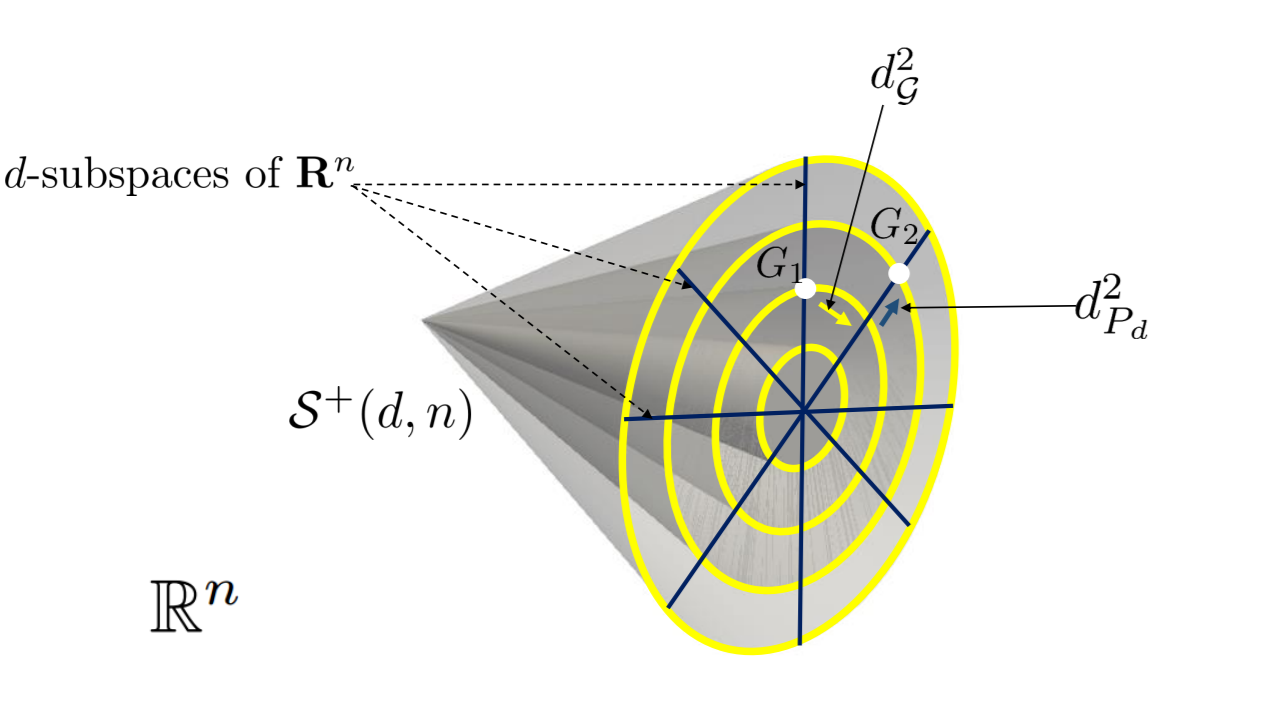}
\caption{A pictorial representation of the positive semidefinite cone $\mathcal{S}^+(d,n)$. Viewing matrices $G_1$ and $G_2$ as ellipsoids in $\mathbb{R}^{n}$; their closeness consists of two contributions: $d^2_{\mathcal{G}}$ (squared Grassmann distance) and $d^2_{\mathcal{P}_d}$ (squared Riemannian distance in $\mathcal{P}_d$).}
\label{Fig:Cone}
\end{figure}

Bonnabel and Sepulchre~\cite{Bonnabel2009SIAM} defined the \textit{pseudo-geodesic} connecting two matrices $G_1 = U_1R_1^2U_1^T$ and $G_2 =  U_2R_2^2U_2^T$ in $\mathcal{S}^+(d,n)$ as the curve
\begin{equation}
\mathcal{C}_{G_1\to G_2}(t)=U(t)R^2(t)U^T(t), \forall t \in [0,1] \; ,
\label{eq:geopsd}
\end{equation}

\noindent where $R^2(t)=R_{1}\exp(t\log R_{1}^{-1}R_{2}^{2}R_{1}^{-1})R_{1}$ is a geodesic in $\mathcal{P}_d$ connecting $R_{1}^2$ and $R_{2}^2$, and $U(t)$ is the geodesic in $\mathcal{G}(d,n)$ given by Eq.~(\ref{eq:GeoGrass}). They also defined the {\it closeness} between $G_1$ and $G_2$, $d_{\mathcal{S}^+}(G_1,G_2)$, as the square of the length of this curve:
\begin{dmath}
\label{eq:closeness}
d_{\mathcal{S}^+}(G_1,G_2)=
d_{\mathcal{G}}^2(\mathcal{U}_{1},\mathcal{U}_{2})+kd_{\mathcal{P}_d}^2(R_{1}^2,R_{2}^2)
=\|\Theta \|_F^2+k\|\log R_{1}^{-1}R_{2}^{2}R_{1}^{-1} \|_F^2 \; ,
\end{dmath}

\noindent where $\mathcal{U}_{i}$ $(i = 1, 2)$ is the $\lspan$ of $U_i$ and $\Theta$ is a $d \times d$ diagonal matrix formed by the principal angles between $\mathcal{U}_{1}$ and $\mathcal{U}_{2}$.
The closeness $d_{\mathcal{S}^+}$ consists of two independent contributions: the square of the distance $d_{\mathcal{G}}(\lspan(U_1),\lspan(U_2))$ between the two associated subspaces, and the square of the distance $d_{\mathcal{P}_d}(R_1^2,R_2^2)$ on the positive cone $\mathcal{P}_d$ (Fig.~\ref{Fig:Cone}). Note that $\mathcal{C}_{G_1\to G_2}$ is not necessarily a geodesic and therefore, the closeness $d_{\mathcal{S}^+}$ is not a true Riemannian distance. From the viewpoint of the landmark configurations $Z_1$ and $Z_2$, with $G_1 = Z_1 Z_1^T$ and $G_2 = Z_2 Z_2^T$, the closeness encodes the distances measured between the affine shapes $\lspan(Z_1)$ and $\lspan(Z_2)$ in $\mathcal{G}(d,n)$ and between their spatial covariances in $\mathcal{P}_d$. Indeed, the spatial covariance of $Z_i$ $(i = 1, 2)$ is the $d \times d$ symmetric positive definite matrix
\begin{equation}
C = \frac{ Z_i^TZ_i}{n-1}=\frac{(U_iR_i)^T(U_iR_i)}{n-1}=\frac{R_i^2}{n-1} \; .
\label{eq:covariance}
\end{equation}

The weight parameter $k$ controls the relative weight of these two contributions. Note that for $k=0$ the distance on $\mathcal{S}^+(d,n)$ collapses to the distance on $\mathcal{G}(d,n)$. Nevertheless, the authors in~\cite{Bonnabel2009SIAM} recommended choosing small values for this parameter. The experiments performed and reported in Section~\ref{sect:results} are in general accordance with this recommendation.

\section{Modeling Temporal Landmark Sequences as Trajectories in $\mathcal{S}^+(d,n)$} \label{sect:temporal-modeling}
We are able to compare static landmark configurations based on their Gramian representation $G$, the induced space, and closeness introduced in the previous Section. We need a natural and effective extension to study their temporal evolution. Following~\cite{Boulbaba2016PAMI,Taheri2001FG,vemulapalli2014human}, we defined curves $\beta_G~:I~\rightarrow~\mathcal{S}^+(d,n)$ ($I$ denotes the time domain, \emph{e.g.,} $[0,1]$) to model the spatio-temporal evolution of elements on $\mathcal{S}^+(d,n)$. Given a sequence of landmark configurations $\{Z_0, \ldots, Z_\tau\}$ represented by their corresponding Gram matrices $\{G_0,\ldots,G_\tau\}$ in $\mathcal{S}^+(d,n)$, the corresponding curve is the trajectory of the point $\beta_G(t)$ on $\mathcal{S}^+(d,n)$, when $t$ ranges in $[0,1]$. These curves are obtained by connecting all successive Gramian representations of shapes $G_i$ and $G_{i+1}$, $0 \leq i\leq \tau-1$, by pseudo-geodesics in $\mathcal{S}^+(d,n)$. Algorithm~\ref{Algo:Trajectory} summarizes the steps to build trajectories in $\mathcal{S}^+(d,n)$ for temporal modeling of landmark sequences.

\begin{algorithm}
\DontPrintSemicolon
\SetKwInOut{Input}{input}\SetKwInOut{Output}{output}
\Input{A sequence of centered landmark configurations $\{Z_0,\cdots, Z_{\tau}\}$, where $Z_{0 \leq i \leq \tau}$ is an $(n \times d)$ matrix ($d=2$ or $d=3$) formed by the coordinates $p_1 = (x_1,y_1),\cdots,p_n = (x_n,y_n)$ or $p_1 = (x_1,y_1,z_1), \cdots, p_n = (x_n,y_n,z_n)$.}
\Output{Trajectory $\beta_G(t)_{0 \leq t \leq \tau}$ and pseudo-geodesics $\mathcal{C}_{\beta_G(t) \to \beta_G(t+1)}$ in $\mathcal{S}^+(d,n)$}
 \textit{ /* Compute the Gram matrices of centered landmarks */ }\\
\For{$i\leftarrow 0$ \KwTo $\tau$}{
  $G_i \longleftarrow Z_iZ_i^T = \langle p_l, p_k\rangle, \; \  1 \leq l, k \leq n$\\
  \textit{/* Compute the Polar decomposition\footnotemark of $Z_i=U_iR_i$ */} \\
  $G_i \longleftarrow U_iR_i^2U_i^T$
 }
 \textit{ /* Compute the pseudo-geodesic paths between successive Gram matrices */ }\\
  $\beta_G(0)\longleftarrow G_0$\\
     \For{$t\leftarrow 0$ \KwTo $\tau-1$}{
$\mathcal{C}_{\beta_G(t) \to \beta_G(t+1)} \longleftarrow \mathcal{C}_{G_{t}\to G_{t+1}}$ given by Eq.~(\ref{eq:geopsd}) connecting $G_{t}$ and $G_{t+1}$ in $\mathcal{S}^+(d,n)$\\
     $\beta_G(t+1)\longleftarrow G_{t+1}$
}
\Return trajectory $\beta_G(t)_{0 \leq t \leq \tau}$ and pseudo-geodesics $\mathcal{C}_{\beta_G(t) \to \beta_G(t+1)}$ in $\mathcal{S}^+(d,n)$
\caption{Computing trajectory $\beta_G(t)$ in $\mathcal{S}^+(d,n)$ of a sequence of landmarks\label{Algo:Trajectory}}
\end{algorithm}
\footnotetext{To compute the polar decomposition, we used the SVD based implementation proposed in~\cite{higham1986computing}.}

\subsection{Temporal Alignment and Rate-Invariant Comparison of Trajectories}\label{sect:DTW}
A relevant issue to our classification problems is -- how to compare trajectories while being invariant to rates of execution? One can formulate the problem of temporal misalignment as comparing trajectories when parameterized differently. The parameterization variability makes the distance between trajectories distorted. This issue was first highlighted by Veeraraghavan~\emph{et al.}~\cite{VeeraraghavanC06} who showed that different rates of execution of the same activity can greatly decrease recognition performance if ignored. Veeraraghan~\emph{et al.}~\cite{VeeraraghavanC06} and Abdelkader~\emph{et al.}~\cite{Abdelkader:2011} used the Dynamic Time Warping (DTW) for temporal alignment before comparing trajectories of shapes of planar curves that represent silhouettes in videos. Following the above-mentioned state-of-the-art solutions, we adopt here a DTW solution to temporally align our trajectories. More formally, given $m$ trajectories $\{\beta_G^{1}, \beta_G^{2}, \ldots, \beta_G^{m} \}$ on $\mathcal{S}^+(d,n)$, we are interested in finding functions $\gamma_i$ such that the $\beta_G^{i}(\gamma_i(t))$ are matched optimally for all $t \in [0,1]$. In other words, two curves $\beta_G^{1}(t)$ and $\beta_G^{2}(t)$ represent the same trajectory if their images are the same. This happens if, and only if, $\beta_G^{2} = \beta_G^{1} \circ \gamma$, where $\gamma$ is a re-parameterization of the interval $[0,1]$. 
The problem of temporal alignment is turned to find an optimal warping function $\gamma^{\star}$ according to,

\begin{equation}
\gamma^{\star}=\arg\min_{\gamma \in \Gamma} \int_0^1 d_{\mathcal{S}^+}(\beta_G^{1}(t),\beta_G^{2}(\gamma(t))) \, \mathrm{d}t \; ,
\label {eq:DTW}
\end{equation}

\noindent where $\Gamma$ denotes the set of all monotonically-increasing functions $\gamma : [0,1] \to [0,1]$. The most commonly used method to solve such optimization problem is DTW. Note that accommodation of the DTW algorithm to the manifold-value sequences can be achieved with respect to an appropriate metric defined on the underlying manifold $\mathcal{S}^+(d,n)$. Having the optimal re-parametrization function $\gamma^{\star}$, one can define a (dis-)similarity measure between two trajectories allowing a rate-invariant comparison:
\begin{equation}
d_{DTW}(\beta_G^{1},\beta_G^{2})=\int_0^1 d_{\mathcal{S}^+}(\beta_G^{1}(t),\beta_G^{2}(\gamma^{\star}(t))) \, \mathrm{d}t \; .
\label {eq:comptraj}
\end{equation}

From now, we shall use $d_{DTW}(.,.)$ to compare trajectories in our manifold of interest $\mathcal{S}^+(d,n)$.

\subsection{Adaptive Re-Sampling of Trajectories in $\mathcal{S}^+(d,n)$}\label{sect:adaptive-res}
One difficulty in video analysis is to capture the most relevant frames and focus on them. In fact, it is relevant to reduce the number of frames when no motion happened, and ``introduce'' new frames, otherwise. Our geometric framework provides tools to do so. In fact, interpolation between successive frames could be achieved using the pseudo-geodesics defined in Eq.~(\ref{eq:geopsd}), while their length (closeness defined in Eq.~(\ref{eq:closeness})) expresses the magnitude of the motion. Accordingly, we have designed an adaptive re-sampling tool that is able to increase/decrease the number of samples in a fixed time interval according to their relevance with respect to the geometry of the underlying manifold $\mathcal{S}^+(d,n)$. Relevant samples are identified by a relatively low closeness $d_{\mathcal{S}^+}$ to the previous frame, while irrelevant ones correspond to a higher closeness level. Here, the down-sampling is performed by removing irrelevant shapes. In turn, the up-sampling is possible by interpolating between successive shape representations in $\mathcal{S}^+(d,n)$, using pseudo-geodesics. 

More formally, given a trajectory $\beta_G(t)_{t=0,1,\ldots,\tau}$ on $\mathcal{S}^+(d,n)$ for each sample $\beta_G(t)$, we compute the closeness to the previous sample, \emph{i.e.,} $d_{\mathcal{S}^+}(\beta_G(t),\beta_G(t-1))$: if the value is below a defined threshold $\zeta_1$, the current sample is simply removed from the trajectory. In contrast, if the distance exceeds a second threshold $\zeta_2$, equally spaced shape representations from the pseudo-geodesic curve connecting $\beta_G(t)$ to $\beta_G(t-1)$ are inserted in the trajectory.

\section{Classification of Trajectories in $\mathcal{S}^+(d,n)$}\label{sect:classification}
Our trajectory representation reduces the problem of landmark sequence classification to that of trajectory classification in $\mathcal{S}^+(d,n)$. That is, let us consider $\mathcal{T}= \{ \beta_G~:[0,1]\rightarrow \mathcal{S}^+(d,n)\}$, the set of time-parameterized trajectories of the underlying manifold. Let $\mathcal{L} = \{(\beta_G^{1},y^{1}),\ldots,(\beta_G^{m},y^{m}) \}$ be the training set with class labels, where $\beta_G^{i} \in \mathcal{T}$ and $y^{i} \in {\cal Y}$, such that $y^i = f(\beta_G^{i})$. The goal here is to find an approximation $h$ to $f$ such that $h: \mathcal{T} \rightarrow \mathcal{L}$. In Euclidean spaces, any standard classifier (\emph{e.g.,} standard SVM) may be a natural and appropriate choice to classify the trajectories. Unfortunately, this is no more suitable in our modeling, as the space $\mathcal{T}$ built from $\mathcal{S}^+(d,n)$ is non-linear. A function that divides the manifold is rather a complicated notion compared with the Euclidean space. In current literature, two main approaches have been used to handle the nonlinearity of Riemannian manifolds~\cite{JayasumanaHSLH15,Taheri2001FG,vemulapalli2014human}. These methods map the points on the manifold to a tangent space or to Hilbert space, where traditional learning techniques can be used for classification. Mapping data to a tangent space only yields a first-order approximation of the data that can be distorted, especially in regions far from the origin of the tangent space. Moreover, iteratively mapping back and forth, \emph{i.e.}, Riemannian Logarithmic and Exponential maps, to the tangent spaces significantly increases the computational cost of the algorithm. Recently, some authors proposed to embed a manifold in a high dimensional Reproducing Kernel Hilbert Space (RKHS), where Euclidean geometry can be applied~\cite{JayasumanaHSLH15}. The Riemannian kernels enable the classifiers to operate in an extrinsic feature space without computing tangent space and $\log$ and $\exp$ maps. Many Euclidean machine learning algorithms can be directly generalized to an RKHS, which is a vector space that possesses an important structure: the inner product. Such an embedding, however, requires a kernel function defined on the manifold which, according to Mercer's theorem, should be positive semi-definite.

\subsection{Pairwise Proximity Function SVM Classifier}
Inspired by a recent work of~\cite{bagheri2016support} for action recognition, we adopted the \textit{pairwise proximity function SVM} (ppfSVM)~\cite{graepel1999classification,gudmundsson2008support}. The ppfSVM requires the definition of a (dis-)similarity measure to compare samples. In our case, it is natural to consider the $d_{DTW}$ defined in Eq.~(\ref{eq:comptraj}) for such a comparison. This strategy involves the construction of inputs such that each trajectory is represented by its (dis-)similarity to all the trajectories, with respect to $d_{DTW}$, in the dataset and then apply a conventional SVM to this transformed data~\cite{gudmundsson2008support}. The ppfSVM is related to the arbitrary kernel-SVM without restrictions on the kernel function~\cite{graepel1999classification}.

Given $m$ trajectories $\{\beta_G^{1}, \beta_G^{2}, \ldots, \beta_G^{m} \}$ in $\mathcal{T}$, following~\cite{bagheri2016support}, a proximity function $\mathcal{P}_{\mathcal{T}}: \mathcal{T} \times \mathcal{T} \rightarrow \mathbb{R}_+$ between two trajectories $\beta_G^{1}, \beta_G^{2} \in \mathcal{T} $ is defined as,
\begin{equation}
\label{eq:proxf}
\mathcal{P}_{\mathcal{T}}(\beta_G^{1}, \beta_G^{2})=d_{DTW}(\beta_G^{1}, \beta_G^{2}) \; .
\end{equation}

According to~\cite{graepel1999classification}, there are no restrictions on the function $\mathcal{P}_{\mathcal{T}}$. For an input trajectory $\beta_G \in \mathcal{T}$, the mapping $\phi(\beta_G)$ is given by,
\begin{equation}
\label{eq:mapping}
\phi(\beta_G)=[\mathcal{P}_{\mathcal{T}}(\beta_G,\beta_G^{1}),\ldots,\mathcal{P}_{\mathcal{T}}(\beta_G,\beta_G^{m}) ]^T \; .
\end{equation}

The obtained vector $\phi(\beta_G) \in \mathbb{R}^m$ is used to represent a sample trajectory $\beta_G \in \mathcal{T}$. Hence, the set of trajectories can be represented by a $m \times m$ matrix $P$, where $P(i,j)=\mathcal{P}_{\mathcal{T}}(\beta_G^{i},\beta_G^{j})$, $i,j \in \{1,\ldots,m\}$. Finally, a linear SVM is applied to this data representation. Further details on ppfSVM can be found in~\cite{bagheri2016support,graepel1999classification,gudmundsson2008support}. In Algorithm~\ref{Algo:Classif}, we provide a pseudo-code for the proposed trajectory classification in $\mathcal{S}^+(d,n)$.

\begin{algorithm}
\DontPrintSemicolon
\SetKwInOut{Input}{input}\SetKwInOut{Output}{output}
\Input{$m$ training trajectories in $\mathcal{S}^+(d,n)$ with their corresponding labels $\{(\beta_G^{1},y^{1}),\ldots,(\beta_G^{m},y^{m}) \}$\\
One testing trajectory $\beta_G^{test}$ in $\mathcal{S}^+(d,n)$}
\Output{Predicted class $y^{test}$ of $\beta_G^{test}$}
 \textit{/* Model training */ }\\
 \For{$i\leftarrow 1$ \KwTo $m$}{
  \For{$j\leftarrow 1$ \KwTo $m$}{
 $P(i,j)=\mathcal{P}_{\mathcal{T}}(\beta_G^{i},\beta_G^{j})$ w.r.t Eq.~(\ref{eq:proxf})
 }
 }
 Training a linear SVM on the data representation $P$\\
 \textit{/* Testing phase */ }\\
$\phi(\beta_G^{test})=[\mathcal{P}_{\mathcal{T}}(\beta_G^{test},\beta_G^{1}),\ldots,\mathcal{P}_{\mathcal{T}}(\beta_G^{test},\beta_G^{m}) ]^T$\\
$y^{test} \longleftarrow $ Linear SVM using the feature vector $\phi(\beta_G^{test})$\\
\Return Predicted class $y^{test}$
\caption{Classification of trajectories in $\mathcal{S}^+(d,n)$   \label{Algo:Classif}}
\end{algorithm}

The proposed ppfSVM classification of trajectories on $\mathcal{S}^+(d,n)$ aims to learn a proximity model of the data, which makes the computation of a pairwise distance function using the DTW (dis-)similarity measure on all the trajectories of the dataset quite necessary. For more efficiency, one can consider faster algorithms for trajectories alignment such us~\cite{salvador2007toward, cuturi2011fast}.

For comparison purposes, we also evaluated a $k$-nearest neighbor solution, where for each test trajectory (sequence), we computed the $k$-nearest trajectories (sequences) from the training set using the same (dis-)similarity measure $d_{DTW}$ defined in Eq.~(\ref{eq:comptraj}). The test sequence is then classified according to a majority voting of its neighbors, (\emph{i.e.}, it is assigned to the class that is most common among its $k$-nearest neighbors).

\section{Experimental Results}\label{sect:results}
To validate the proposed framework, we conducted extensive experiments on three human behavior understanding applications. These scenarios show the potential of the proposed solution when landmarks capture different information on different data.
First, we addressed the problem of activity recognition from depth sensors such as the Microsoft Kinect. In this case, $3$D landmarks correspond to the joints of the body skeleton, as extracted from RGB-Depth frames. The number of joints per skeleton varies between $15$ and $20$, and their position is generally noisy. Next, we addressed the new emerging problem of finding relationships between body movement and emotions using $3$D skeletal data. Here, landmarks correspond to physical markers placed on the body and tracked with high temporal rate and good estimation of the $3$D position by a Motion Capture (MoCap) system. 
Finally, we evaluated our framework on the problem of facial expression recognition using landmarks of the face. In this case, $49$ face landmarks are extracted in $2$D with high accuracy using a state-of-the-art face landmark detector.

\subsection{3D Human Action Recognition}\label{sect:action-recognition}
Action recognition has been performed on $3$D skeleton data as provided by a Kinect camera in different datasets. In this case, landmarks correspond to the estimated position of $3$D joints of the skeleton ($d$=3). With this assumption, skeletons are represented by $n \times n$ Gram matrices of rank $3$ lying on $\mathcal{S}^{+}(3,n)$, and skeletal sequences are seen as trajectories on this manifold.

As discussed in Section~\ref{sect:representation}, the information given by the Gram matrix of the skeleton is linearly equivalent to that of the pairwise distances between different joints. Thus, considering only some specific subparts of the skeletons can be more accurate for some actions. For instance, it is more discriminative to consider only the pairwise distances between the joints of left and right arms for actions that involve principally the motion of arms, (\emph{e.g.}, \textit{wave hands}, \textit{throw}). Accordingly, we divided the skeletons into three body parts, \emph{i.e.}, left/right arms, left/right legs and torso, while keeping a coarse information given by all the joints of the skeleton (we show an example of this decomposition in the supplementary material). For an efficient use of the information given by the different body parts, we propose a late fusion of four ppf-SVM classifiers that consists of: (1) training all the body part classifiers separately; (2) merging the contributions of the four body part classifiers. This is done by multiplying the probabilities $s_{i,j}$, output of the SVM for each class $j$, where $i \in \{1,2,3,4\}$ denotes the body part. The class $\mathcal{C}$ of each test sample is determined by
\begin{equation}
\label{eq:fusion}
\mathcal{C}=\underset{j}{\operatorname*{arg\,max}}\prod_{i=1}^{4} s_{i,j}, \; j=1, \dots ,n_{\mathcal{C}} \; ,
\end{equation}

\noindent where $n_{\mathcal{C}}$ is the number of classes.

\begin{table*}[ht]
  \centering
  \caption{Overall accuracy (\%) on the UT-Kinect, Florence3D, SBU interaction, and SYSU-3D datasets. Here, $^{(D)}$: depth; $^{(C)}$: color (or RGB); $^{(G)}$: geometry (or skeleton); $^*$: Deep Learning based approach; last row: ours}\label{tab:ARR_Action}
  \footnotesize
\begin{tabular}{{l}||{c}|{c}||{c}|{c}||{c}|{c}||{c}|{c}}
  & \multicolumn{2}{c||}{\textbf{UT-Kinect}} & \multicolumn{2}{c||}{\textbf{Florence3D}}  & \multicolumn{2}{c||}{\textbf{SBU Interaction}} & \multicolumn{2}{c}{\textbf{SYSU-3D}} \\
 \hline
  \textbf{Method} & \textbf{Protocol} & \textbf{Acc (\%)}  &  \textbf{Protocol} & \textbf{Acc (\%)} &  \textbf{Protocol} & \textbf{Acc (\%)}  &  \textbf{Protocol} & \textbf{Acc (\%)} \\
  \hline
    $^{(G+D)}$ 3D$^2$CNN \cite{liu20163d}$^*$ &  LOSO  & 95.5 & -- & -- & --& --& --& --\\
     $^{(G+D+C)}$ Dynamic features \cite{hu2015jointly}  & --  &  -- & --  & -- & -- & --& Half-Half& \textbf{84.9  $\pm$ 2.29}\\
    $^{(G+D+C)}$ LAFF \cite{hu2016real} &  --  & -- & -- & --& --& --&Half-Half& 80\\
    \hline

     $^{(G)}$  LARP \cite{vemulapalli2014human} &  $5$-fold & 97.08 & $5$-fold & 90.88 & -- & --& --& --\\
     
      $^{(G)}$ Gram Hankel \cite{Zhang_2016_CVPR}  &  LOOCV  & \textbf{100} & -- & -- & -- & --& --& --\\
        $^{(G)}$  Motion trajectories \cite{Devanne2015Cybernetics} &  LOOCV  & 91.5 & LOSO & 87.04 & -- & --& --& --\\
     $^{(G)}$  Elastic func. coding \cite{anirudh:2017} &  $5$-fold  & 94.87 & $5$-fold & 89.67 & -- & --& --& --\\

      $^{(G)}$ Mining key poses \cite{wang2016mining} &  LOOCV & 93.47 & LOSO & \textbf{92.25} & --& --& --& --\\
      $^{(G)}$ NBNN+parts+time \cite{seidenari2013recognizing} &  -- & -- & LOSO & 82 & --& --& --& --\\
    $^{(G)}$ LAFF (SKL) \cite{hu2016real} &  --  & -- & -- & --& --& --&Half-Half& 54.2\\
    $^{(G)}$ Dynamic skeletons \cite{hu2015jointly}  & --  &  -- & --  & -- & -- & --&Half-Half& 75.5 $\pm$ 3.08\\
    $^{(G)}$ LSTM-trust gate \cite{Liu2016}$^*$  &  LOOCV  & 97.0 & -- &-- &$5$-fold& 93.3&Half-Half&\textbf{76.5}\\
    $^{(G)}$ JL-distance LSTM\cite{zhang2017geometric}$^*$ & $5$-fold &  95.96  & --  & -- & $5$-fold & \textbf{99.02} & --& --\\
    $^{(G)}$ Co-occurence LSTM\cite{zhu:2016}$^*$   & -- &  --  & -- & -- & $5$-fold & 90.41& --& --\\
     $^{(G)}$ Hierarchical RNN\cite{du:2015}$^*$   & -- &  --  & -- & -- & $5$-fold & 80.35& --& --\\
     $^{(G)}$ SkeletonNet\cite{ke2017skeletonnet}$^*$   & -- &  --  & -- & -- & $5$-fold & 93.47& --& --\\
     $^{(G)}$ STA-LSTM\cite{song2017end}$^*$   & -- &  --  & -- & -- & $5$-fold & 91.51& --& --\\

        \hline
     Traj. on $\mathcal{G}(3,n)$ (full body)  &  LOOCV  &  92.46 & LOSO &75 $\pm$ 5.22& $5$-fold  & 76.3 $ \pm$ 3.26&  Half-Half& 73.26 $\pm$ 2.27\\

     Traj. on $\mathcal{G}(3,n)$ - BP Fusion & LOOCV  &  96.48 & LOSO &  76.4 $\pm$ 5.37&$5$-fold&83.56 $\pm$ 4.72& Half-Half& 76.61 $\pm$ 2.86 \\

     \textbf{Traj. on $\mathcal{S}^{+}(3,n)$ (full body)}  &  LOOCV  & \textbf{96.48} & LOSO  & \textbf{88.07$ \pm$ 4.8} & $5$-fold& \textbf{88.45 $\pm$ 2.88}& Half-Half & \textbf{76.01 $\pm$ 2.09} \\

    \textbf{Traj. on $\mathcal{S}^{+}(3,n)$ - BP Fusion} &  LOOCV  & \textbf{98.49} & LOSO &\textbf{88.85 $\pm$ 4.6}& $5$-fold & \textbf{ 93.7 $\pm$ 1.59} & Half-Half& \textbf{80.22$ \pm$ 2.09}  \\

  \hline
  \end{tabular}
\end{table*}

\subsubsection{Datasets}
We performed experiments on four publicly available datasets showing different challenges. All these datasets have been collected with a Microsoft Kinect sensor. 

\textbf{UT-Kinect dataset}~\cite{xia2012view} -- It contains $10$ actions performed by $10$ different subjects. Each subject performed each action twice resulting in $199$ valid action sequences. The 3D locations of $20$ joints are provided with the dataset.

\textbf{Florence3D dataset}~\cite{seidenari2013recognizing} -- It contains $9$ actions performed two or three times by $10$ different subjects. Skeleton comprises $15$ joints. This is a challenging dataset due to variations in the view-point and large intra-class variations.

\textbf{SYSU-3D dataset}~\cite{hu2015jointly} --  It contains $480$ sequences. In this dataset, $12$ different activities focusing on interactions with objects were performed by $40$ persons. The $3$D coordinates of $20$ joints are provided in this dataset. The SYSU-3D dataset is very challenging since the motion patterns are highly similar among different activities.

\textbf{SBU Interaction dataset}~\cite{yun2012two} -- This dataset includes $282$ skeleton sequences of eight types of two-persons interacting with each other, including \textit{approaching}, \textit{departing}, \textit{pushing}, \textit{kicking}, \textit{punching}, \textit{exchanging objects}, \textit{hugging}, and \textit{shaking hands}. In most interactions, one subject is acting, while the other subject is reacting.

\subsubsection{Experimental Settings and Parameters}\label{sect:evalsettings}
For all the datasets, we used only the provided skeletons. The adaptive re-sampling of trajectories discussed in Section~\ref{sect:adaptive-res} has been not applied on these data. The motivation is that this operation tries to capture small shape deformations of the landmarks and this can amplify the noise of skeleton joints. For the SBU dataset, where two skeletons of two interacting persons are given in each frame, we considered all the joints of the two skeletons. In this case, a unique Gram matrix is computed for the two skeletons modeling the interaction between them. In this dataset, the decomposition into body parts is performed only for the acting person since the other person is reacting in a coarse manner.

As discussed in Section~\ref{sect:pseudo-geodesics}, our body movement representation involves a parameter $k$ that controls the contribution of two information: the affine shape of the skeleton at time $t$, and its spatial covariance. The affine shape information is given by the Grassmann manifold $\mathcal{G}(3,n)$, while the spatial covariance is given by the SPD manifold $\mathcal{P}_3$. We recall that for $k=0$, the skeletons are considered as trajectories on the Grassmann manifold $\mathcal{G}(3,n)$. For each dataset, we performed a cross-validation grid search, $k \in [0,3]$ with a step of $0.1$, to find an optimal value $k^*$. In the case of skeleton decomposition into body parts, a different parameter $k$ is used for computing the distance of each body part, (\emph{i.e.}, one parameter each for arms, legs, and torso, and one parameter for the whole skeleton). Each parameter $k$ is evaluated separately by a cross-validation grid search in the classifier of the relative body part.

To allow a fair comparison, we adopted the most common experimental settings in literature. For the UT-Kinect dataset, we used the \textit{leave-one-out cross-validation} (LOOCV) protocol~\cite{xia2012view}, where one sequence is used for testing and the remaining sequences are used for training. For the Florence3D dataset, a \textit{leave-one-subject-out} (LOSO) schema is adopted following~\cite{Devanne2015Cybernetics,wang2016mining,Zhang_2016_CVPR}. For the SYSU3D dataset, we followed~\cite{hu2015jointly} and performed a \textit{Half-Half} cross-subject test setting, in which half of the subjects were used for training and the remaining half were used for testing. Finally, a \textit{$5$-fold} cross-validation was used for the SBU dataset. Note that the subjects considered in each split are those given by the datasets (SYSU3D and SBU). All our programs were implemented in Matlab and run on a $2.8$ GHZ CPU. We used the multi-class SVM implementation of the LibSVM library~\cite{chang2011libsvm}.

\subsubsection{Results and Discussion}\label{sect:action-recognition-results}
In Table~\ref{tab:ARR_Action}, we compare our approach with existing methods dealing with skeletons and/or RGB-D data. Overall, our approach achieved competitive results compared to recent state-of-the-art approaches. We provide the confusion matrices for all the datasets in the supplementary material.

On the UT-Kinect dataset, we obtained an average accuracy of $96.48\%$, when considering the full skeletal shape. Using a late fusion of classifiers based on the body parts, as described in Section~\ref{sect:action-recognition}, the performance increased to $98.49\%$ outperforming~\cite{Liu2016,Devanne2015Cybernetics,wang2016mining}. The highest average accuracy for this dataset was reported in~\cite{Zhang_2016_CVPR} ($100\%$), where Gram matrices were used for skeletal sequence representation, but in a completely different context. Specifically, the authors of~\cite{Zhang_2016_CVPR} built a Gram matrix from the Hankel matrix of an Auto-Regressive (AR) model that represented the dynamics of the skeletal sequences.
The used metric for the comparison of Gram matrices is also different than ours as they used metrics in the positive definite cone by regularizing their ranks, \emph{i.e.}, making them full-rank.

On the SBU dataset, the fusion of body parts achieved the highest accuracy reaching $93.7\%$. We observed that all the interactions present in this dataset are well recognized, \emph{e.g.}, \textit{hugging} ($100\%$), \textit{approaching} ($97.5\%$), etc., except \textit{pushing} ($74.7\%$), which has been mainly confused with a very similar interaction, \emph{i.e.}, \textit{punching}. Here, our approach is ranked second after~\cite{zhang2017geometric}, where an average accuracy of $99.02\%$ is reported. In that work, the authors compute a large number of joint-line distances per frame making their approach time consuming.

On the SYSU3D dataset, our approach achieved the best result compared to skeleton based approaches. We report an average accuracy of $80.22\%$ with a standard deviation of $2.09\%$, when the late fusion of body parts is used. Our approach, applied to the full skeleton, still achieved very competitive results and reached $76.01\%$ with a standard deviation of $2.09\%$. Combining the skeletons with depth and color information, including the object, Hu~\emph{et al.}~\cite{hu2015jointly} obtained the highest performance with an average accuracy of $84.9\%$ and a standard deviation of $2.29\%$.

On the Florence3D dataset, we obtained an average accuracy of $88.07\%$, improved by around $0.8\%$ when involving body parts fusion. While high accuracies are reported for coarse actions, \emph{e.g.}, \textit{sitting down} ($95\%$), \textit{standing up} ($100\%$), and \textit{lacing} ($96.2\%$), finer actions, \emph{e.g.}, \textit{reading watch} ($73.9\%$) and \textit{answering phone} ($68.2\%$) are still challenging. Our results are outperformed by~\cite{wang2016mining,vemulapalli2014human}, where the average accuracies are greater than $90\%$.

\textbf{Baseline Experiments}. 
In this paragraph, we discuss the effect of using the different steps in our framework and their computational complexity compared to baselines. Results of this evaluation are reported in Table~\ref{tab:baseline_action}. Firstly, in the top part of Table~\ref{tab:baseline_action}, we studied the computational cost of the proposed pipeline in the task of 3D action recognition and report running time statistics for the different steps of our approach on UT-Kinect dataset. Specifically, we provide the necessary execution time for: (1) an arbitrary trajectory construction in $\mathcal{S}^+(3,n)$ as described in Algorithm~\ref{Algo:Trajectory}; (2) comparison of two arbitrary trajectories with the proposed version of DTW; (3) testing phase of an arbitrary trajectory classification with ppfSVM in $\mathcal{S}^+(3,n)$ as described in Algorithm~\ref{Algo:Classif}.

\begin{table}[ht]
  \centering
  \caption{Baseline experiments on the UT-Kinect, SBU, SYSU3D, and Florence3D datasets}\label{tab:baseline_action}
  \footnotesize

\begin{tabular}{{l}|{c}}
\textbf{Pipeline component} & \textbf{Time} (s) \\
        \hline 
    Trajectory construction in $\mathcal{S}^+(3,n)$& 0.007 \\
    Comparison of trajectories in $\mathcal{S}^+(3,n)$& 0.93 \\
    Classification of a trajectory in $\mathcal{S}^+(3,n)$& 147.71 \\
    \hline
  \end{tabular}

  \vspace{0.3cm}
\begin{tabular}{{l}|{c}|{c}}
\textbf{Distance} & \textbf{UT-Kinect} (\%) & \textbf{Time} (s) \\
        \hline 
    Flat distance $d_{\mathcal{F}^+}$ & 92.96 & 0.06 \\
    Distance $d_{\mathcal{P}_n}$ in $\mathcal{P}_n$  & 94.98 & 1.66 \\
    Closeness $d_{\mathcal{S}^+}$ & \textbf{96.48} & 0.93 \\
    \hline
  \end{tabular}

\vspace{0.3cm}

  \begin{tabular}{{l}|{c}|{c}|{c}}
\textbf{Temp. alignment} & \textbf{UT-Kinect (\%)} & \textbf{SBU (\%)} & \textbf{Time} (s) \\
        \hline
    No DTW & 91.46 & 81.36$ \pm$ 2.78 & 0.02 \\
    DTW &  \textbf{96.48} &\textbf{88.45$ \pm$ 2.88} & 0.93 \\
    \hline
  \end{tabular}

  \vspace{0.3cm}

\begin{tabular}{{l}|{c}|{c}}
\textbf{Classifier} & \textbf{UT-Kinect (\%)} & \textbf{SBU (\%)} \\
      \hline 
  K-NN --  $\mathcal{G}(3,n)$ & 86.93 & 42.72 $\pm$ 5.68 \\
 Ppf-SVM --  $\mathcal{G}(3,n)$ & 92.46 & 76.3 $\pm$ 3.26 \\
  K-NN -- $\mathcal{S}^{+}(3,n)$ & 91.96 & 61.06 $\pm$ 2.3 \\
Ppf-SVM --   $\mathcal{S}^{+}(3,n)$  & \textbf{96.48} &  \textbf{88.45$ \pm$ 2.88} \\
  \hline
 \end{tabular}

  \vspace{0.3cm}

 \begin{tabular}{{l}|{c}|{c}}
\textbf{Body parts} & \textbf{UT-Kinect (\%)} & \textbf{SBU (\%)} \\
      \hline 
  Arms only & 87.94 & 80.96 $\pm$ 5.53 \\
  Legs only & 35.68 & 83.36 $\pm$ 2.41 \\
 Torso only & 72.36 & 80.58 $\pm$ 2.16\\
Whole body & 96.48 &  88.45 $\pm$ 2.88 \\
Late BP Fusion & \textbf{98.49} &  \textbf{93.7 $\pm$ 1.59} \\
  \hline
 \end{tabular}

 \vspace{0.3cm}

  \begin{tabular}{{l}|{c}|{c}}
\textbf{Body parts} & \textbf{Florence3D (\%)} & \textbf{SYSU3D (\%)} \\
      \hline 
  Arms only & 75.72 $\pm$ 8.45 & 73.88 $\pm$ 2.64 \\
  Legs only & 42.44 $\pm$ 7.69 & 37.6 $\pm$ 2.10 \\
 Torso only & 54.33 $\pm$ 10.62 &49.36 $\pm$ 3.94 \\
Whole body & 88.07 $\pm$ 4.8 & 76.01$\pm$ 2.09 \\
Late BP Fusion & \textbf{88.85 $\pm$ 4.6} & \textbf{80.22 $\pm$ 2.09} \\
  \hline
 \end{tabular}
\end{table}

Secondly, we can observe the large superiority of the Gramian representation over the Grassmann representation. For the Florence3D and SBU datasets, we report an improvement of about $12\%$. For UT-Kinect and SYSU3D, the performance increased by about $3\%$. Note that these improvements over the Grassmannian representation are due to the additional information of the spatial covariance given by the SPD manifold in the metric. The contribution of the spatial covariance is weighted with a parameter $k$. As discussed in Section~\ref{sect:evalsettings}, we performed a grid search cross-validation to find the optimal value $k^*$ of this parameter. The optimal values are $k^*= 0.05$, $k^*=0.81$, $k^*=0.25$, and $k^*=0.09$ for the the UT-Kinect, SBU, Florence3D, and SYSU3D datasets, respectively. These results are in concordance with the recommendation of Bonnabel and Sepulchre~\cite{Bonnabel2009SIAM} to use relative small values of $k$.

Then, we evaluated the proposed metric with respect to other metrics used in state of the art solutions. Specifically, given two matrices $G_1$ and $G_2$ in $\mathcal{S}^+(3,n)$, we compared our results with two other possible metrics: (1) as proposed in~\cite{wang2012covariance,Zhang_2016_CVPR}, we used $d_{\mathcal{P}_n}$ that was defined in Eq.~(\ref{eq:closeness}) to compare $G_1$ and $G_2$ by regularizing their ranks, \emph{i.e.}, making them $n$ full-rank, and considering them in $\mathcal{P}_n$ (the space of $n$-by-$n$ positive definite matrices), $d_{\mathcal{P}_n}(G_1,G_2)=d_{\mathcal{P}_n}(G_1+\epsilon I_n,G_2+\epsilon I_n)$; (2) we used the Euclidean flat distance $d_{\mathcal{F}^+}(G_1,G_2)=\|G_1-G_2\|_F$, where $\|.\|_F$ denotes the Frobenius-norm. Note that the provided execution times are relative to the comparison of two arbitrary sequences. We can observe that in Table~\ref{tab:baseline_action}, the closeness $d_{\mathcal{S}^+}$ between two elements of $\mathcal{S}^+(3,n)$ defined in Eq.~(\ref{eq:closeness}) is more suitable compared to the distance $d_{\mathcal{P}_n}$ and the flat distance $d_{\mathcal{F}^+}$ defined in literature. This demonstrates the importance of considering the geometry of the manifold of interest. Another advantage of using $d_{\mathcal{S}^+}$ over $d_{\mathcal{P}_n}$ is the computational time as it involves $n$-by-$3$ and $3$-by-$3$ matrices instead of $n$-by-$n$ matrices.

To show the relevance of aligning the skeleton sequences in time before comparing them, we conducted the same experiments without using Dynamic Time Warping (DTW). In this case, the performance decreased by around $5\%$ and $7\%$ on UT-Kinect and SBU datasets, respectively. Here, the provided execution times are relative to the comparison of two arbitrary sequences on UT-Kinect dataset.
Furthermore, we also compared the proposed ppfSVM classifier with a $k$-nearest neighbor classifier.
The number of nearest neighbors $k$ to consider for each dataset is chosen by cross-validation. Using the $k$-NN classifier, we obtained an average accuracy of $91.96\%$ with $k=5$ neighbors on UT-Kinect and $61.06\%$ with $k=4$ on the SBU dataset. These results are outperformed by the ppfSVM classifier.

Finally, in Table~\ref{tab:baseline_action} we provide the obtained accuracies when considering the different body parts separately on all the datasets. Unsurprisingly, the highest accuracy is achieved by left and right arms in all the datasets compared to the torso and the legs, since the majority of the actions are acted using arms. One can note the considerable improvements realized by the late fusion compared to the whole skeleton in all the datasets, especially in the SBU and SYSU3D datasets, where we report improvements of about $5\%$ and $4\%$, respectively.

\subsection{Emotion Recognition from 3D Body Movement}
Recently, the study of computational models for human emotion recognition has gained increasing attention not only for commercial applications (to get feedback on the effectiveness of advertising material), but also for gaming and monitoring of the emotional state of operators that act in risky contexts such as aviation. Most of these studies have focused on the analysis of facial expressions, but important clues can be derived by the analysis of the dynamics of body parts as well~\cite{hicheur:2010}.
Using the same geometric framework that was proposed for action recognition, we evaluated our approach in the task of emotion recognition from human body movement. Here, the used landmarks are in $3$D coordinate space, but with better accuracy and higher temporal resolution, with respect to the case of action recognition.

\subsubsection{Dataset}
Experiments have been performed on the Body Motion-Emotion dataset (P-BME), acquired at the Cognitive Neuroscience Laboratory (INSERM U960 - Ecole Normale Sup\'{e}rieure) in Paris~\cite{hicheur:2010}.
It includes Motion Capture (MoCap) $3$D data sequences recorded at high frame rate ($120$ frames per second) by an Opto-electronic Vicon V8 MoCap system wired to $24$ cameras.
The body movement is captured by using $43$ landmarks that are positioned at joints.

To create the dataset, $8$ subjects (professional actors) were instructed to walk following a predefined ``U'' shaped path that includes forward-walking, turn, and coming back. For each acquisition, actors moved along the path performing one emotion out of five different emotions, namely, \textit{anger}, \textit{fear}, \textit{joy}, \textit{neutral}, and \textit{sadness}. So, each sequence is associated with one emotion label. 
Each actor performed at maximum five repetitions of a same emotional sequence for a total of $156$ instances.
Though there is some variation from subject to subject, the number of examples is well distributed across the different emotions: $29$ \textit{anger}, $31$ \textit{fear}, $33$ \textit{joy}, $28$ \textit{neutral}, $35$ \textit{sadness}.

\subsubsection{Experimental Settings and Parameters}
Since MoCap skeletons are in $3$D coordinate space, we followed the same steps that have been proposed for action recognition, including the decomposition into body parts (we show an example of this decomposition in the supplementary material). Note that the same late fusion of body part classifiers, as mentioned in the previous Section, is adopted. A cross-validation grid search has been performed to find an optimal value for the weight parameter $k$.

Experiments on the P-BME dataset were performed by using a \textit{leave-one-subject-out} cross validation protocol. With this solution, iteratively, all the emotion sequences of a subject are used for test, while all the sequences of the remaining subjects are used for training.

\subsubsection{Results and Discussion}

In Table~\ref{tab:ARR_BodyEmo}, we provide the obtained results as well as a comparative study with baseline experiments on the P-BME dataset.

\begin{table}[ht]
\centering
\caption{Comparative study of the proposed approach with baseline experiments on the P-BME dataset. First rows: state-of-the-art action and emotion recognition methods and human evaluator; second rows: baseline experiments; last row: ours}\label{tab:ARR_BodyEmo}
  \footnotesize
\begin{tabular}{{l}|{c}}
\textbf{Method} & \textbf{ Accuracy (\%)}\\
    \hline
    Human evaluator & 74.20 \\
    COV3D~\cite{daoudi2017emotion} & 71.14 $\pm$ 6.77\\
    LARP~\cite{vemulapalli2014human} & 74.8 $\pm$ 3.17\\
     \hline
    Traj. on $\mathcal{S}^+(3,n)$ - Flat metric & 57.41 $\pm$ 8.43\\
    Traj. on $\mathcal{S}^+(3,n)$ - No DTW & 63.23 $\pm$ 8.62\\
    Traj. on $\mathcal{S}^+(3,n)$ - $k$NN & 68.9 $\pm$ 7.63\\
    \hline
    Traj. on $\mathcal{G}(3,n)$ & 66.35$ \pm$ 6.43 \\
    Traj. on $\mathcal{G}(3,n)$ - BP Fusion & 67.09 $\pm$ 6.82 \\
    \textbf{Traj. on $\mathcal{S}^+(3,n)$} & \textbf{78.15 $\pm$ 5.79 } \\
    \textbf{Traj. on $\mathcal{S}^+(3,n)$} - BP Fusion & \textbf{81.99 $\pm$ 4.36 } \\
     \hline
\end{tabular}
\end{table}

Similarly to the reported results for action recognition, the proposed fusion of body part classifiers achieved the highest performance with an average accuracy of $81.99\%$ and standard deviation of $4.36\%$. The best results were scored by \textit{neutral} and \textit{anger} (more than $80\%$), followed by \textit{fear} ($71\%$), \textit{joy} (about $67\%$), with the lowest accuracy for \textit{sadness} (about 65\%). In the supplementary material, we provide the related confusion matrix as well as some additional experiments.
Considering only the skeletons (without body parts) in the classification, the performance decreased to an average accuracy of $78.15\%$.

Recently, Daoudi~\emph{et al.}~\cite{daoudi2017emotion} proposed a method for emotion recognition from body movement based on covariance matrices and SPD manifold. They used the $3$D covariance descriptor (COV3D) of skeleton joints across time to represent sequences without a special handling of the dynamics. They reported and average accuracy of $71.4\%$. They also performed a user based test in order to evaluate the performance of the proposed classification method in comparison with a human-based judgment. In this test, thirty-two naive individuals were asked to perform a force-choice task in which they had to choose between one of the five emotions. This resulted in an average value of about $74\%$. It is relevant to note that the user based test being based on RGB videos provides to the users much more information for evaluation, including the actor's face. Notably, our method is capable to score better results based on the skeleton joints only.

We also compared our results with the Lie algebra relative pairs (LARP) method proposed by Vemulapalli~\emph{et al.}~\cite{vemulapalli2014human} for skeleton action recognition. In that work, each skeleton is mapped to a point on the product space of $SE(3) \times SE(3) \cdots \times SE(3)$, where it is modeled using transformations between joint pairs. The temporal evolution of these features is seen as a trajectory on $SE(3) \times SE(3) \times \cdots \times SE(3)$ and mapped to the tangent space of a reference point. A one-versus-all SVM combined with Dynamic Time Warping and Fourier temporal pyramid (FTP) is used for classification. Using this method, an average accuracy of $74.8\%$ was obtained, which is about $8\%$ lower than ours.

The highest accuracy ($78.15\%$) is obtained for $k^*=1.2$. For $k=0$, the skeletons are considered as trajectories on the Grassmann manifold $\mathcal{G}(3,n)$, and the obtained accuracy is around $66\%$, which is $12\%$ lower than the retained result. 
In order to show the importance of choosing a well defined Riemannian metric in the space of interest, we conducted the same experiments by changing the metric $d_{\mathcal{S}^+}$ defined in Eq.~\eqref{eq:closeness} with a flat metric, defined as the Frobenius norm of the difference between two Gram matrices (skeletons). For this experiment, we report an average accuracy of $57.41\%$ being lower of about $21\%$ than using $d_{\mathcal{S}^+}$.

Finally, as mentioned in Section~\ref{sect:DTW}, an important step in our approach is the temporal alignment. Avoiding this step and following the same protocol, we found that the performance decreased to $63.23\%$.
We also studied the method when considering the different body parts separately and also when considering different sequence lengths. Results of these additional experiments are reported in the supplementary material.

\subsection{2D Facial Expression Recognition}\label{sect:facial-expressions}
We evaluated our approach also in the task of facial expression recognition from $2$D landmarks. In this case, the landmarks are in a $2$D coordinate space, resulting in a Gram matrix of size $n \times n$ of rank $2$ for each configuration of $n$ landmarks. The facial sequences are then seen as time-parameterized trajectories on $\mathcal{S}^{+}(2,n)$.

\subsubsection{Datasets}
We conducted experiments on four publicly available datasets -- CK+, MMI, Oulu-CASIA, and AFEW datasets.

\textbf{Cohn-Kanade Extended (CK+) dataset}~\cite{LuceyCKSAM10} -- 
It contains $123$ subjects and $593$ frontal image sequences of posed expressions. Among them, $118$ subjects are annotated with the seven labels -- \textit{anger} (An), \textit{contempt} (Co), \textit{disgust} (Di), \textit{fear} (Fe), \textit{happy} (Ha), \textit{sad} (Sa) and \textit{surprise} (Su). Note that only the two first temporal phases of the expression, \emph{i.e.}, neutral and onset (with apex frames), are present.

\textbf{MMI dataset}~\cite{Valstar2010idhas} -- It consists of $205$ image sequences with frontal faces of $30$ subjects labeled with the six basic emotion labels. In this dataset each sequence begins with a neutral facial expression, and has a posed facial expression in the middle; the sequence ends up with the neutral facial expression. The location of the peak frame is not provided as a prior information.

\textbf{Oulu-CASIA dataset}~\cite{ZhaoHTLP11} -- It includes $480$ image sequences of $80$ subjects, taken under normal illumination conditions. They are labeled with one of the six basic emotion labels. Each sequence begins with a neutral facial expression and ends with the apex of the expression.

\textbf{AFEW dataset}~\cite{DhallGLG12} -- Collected from movies showing close-to-real-world conditions, which depict or simulate the spontaneous expressions in uncontrolled environment. The task is to classify each video clip into one of the seven expression categories (the six basic emotions plus the neutral).

\subsubsection{Experimental Settings and Parameters}
All our experiments were performed once facial landmarks were extracted using the method proposed in~\cite{AsthanaZCP14} on the CK+, MMI, and Oulu-CASIA datasets. On the challenging AFEW dataset, we have considered the corrections provided in\footnote{ http://sites.google.com/site/chehrahome} after applying the same detector. The number of landmarks is $n=49$ for each face.
In this case, we applied the adaptive re-sampling of trajectories proposed in Section~\ref{sect:adaptive-res} that enhances small facial deformations and disregards redundant frames. This step involves two parameters $\zeta_1$ and $\zeta_2$ for up-sampling and down-sampling, respectively. These two parameters are chosen so that all the trajectories in the dataset have the same length, equal to the median length. For the parameter $k$, the same procedure as for action and emotion recognition from body movement is applied.

To evaluate our approach, we followed the experimental settings commonly used in recent works. Following~\cite{ElaiwatBB16,JungLYPK15,LiuSWC14,ZhongLYLHM12}, we have performed $10$-fold cross validation experiments for the CK+, MMI, and Oulu-CASIA datasets. In contrast, the AFEW dataset was divided into three sets: training, validation and test, according to the protocols defined in EmotiW'2013~\cite{DhallGJWG13}. Here, we only report our results on the validation set for comparison with~\cite{DhallGJWG13,ElaiwatBB16,LiuSWC14}.

\subsubsection{Results and Discussion}
On CK+, the average accuracy is $96.87\%$. Note that the accuracy of the trajectory representation on $\mathcal{G}(2,n)$, following the same pipeline is $2\%$ lower, which confirms the contribution of the covariance embedded in our representation.

An average classification accuracy of $79.19\%$ is reported for the MMI dataset. Note that based on geometric features only, our approach grounding on both representations on $\mathcal{S}^+(2,n)$ and $\mathcal{G}(2,n)$ achieved competitive results with respect to the literature (see Table~\ref{tab:ARR_CK+_MMI}).
On the Oulu-CASIA dataset, the average accuracy is $83.13\%$, hence $3\%$ higher than the Grassmann trajectory representation. This is the highest accuracy reported in literature (refer to Table~\ref{tab:ARR_AFEW}).
Finally, we reported an average accuracy of $39.94\%$ on the AFEW dataset. Despite being competitive with respect to recent literature (see Table~\ref{tab:ARR_AFEW}), these results evidence that AFER "in-the-wild" is still challenging.

We highlight the superiority of the trajectory representation on $\mathcal{S}^+(2,n)$ over the Grassmannian (refer to Table~\ref{tab:ARR_CK+_MMI} and Table~\ref{tab:ARR_AFEW}). This is due to the contribution of the covariance part further to the conventional affine-shape analysis over the Grassmannian. Recall that $k$ serves to balance the contribution of the distance between covariance matrices living in $\mathcal{P}_2$ with respect to the Grassmann contribution $\mathcal{G}(2,n)$. The optimal performance are achieved for the following values -- $k^*_{CK+}=0.081$, $k^*_{MMI}=0.012$, $k^*_{Oulu-CASIA}=0.014$ and $k^*_{AFEW}=0.001$. To show the importance of the proposed adaptive re-sampling step, we conducted the same experiments on the MMI and AFEW datasets avoiding this step. The performance decreases of about $5\%$ on MMI and $3\%$ on AFEW.
The temporal alignment, the effectiveness of the used metric, and the used classifier were also evaluated according to the same conducted baseline experiments for action recognition (see Section~\ref{sect:action-recognition-results}). Results show the superiority of the retained framework over these baselines. Further details, as well as the confusion matrices, are available in the supplementary material of this paper.

\textbf{Comparative Study with the State-of-the-Art}.\label{sec:comp} 
In Table~\ref{tab:ARR_CK+_MMI} and Table~\ref{tab:ARR_AFEW}, we compare our approach over the recent literature. Overall, our approach achieved competitive performance with respect to the most recent approaches. On CK+, we obtained the second highest accuracy. The ranked-first approach is DTAGN~\cite{JungLYPK15}, in which two deep networks are trained on shape and appearance channels, then fused. Note that the geometry deep network (DTGN) achieved $92.35\%$, which is much lower than ours. Furthermore, our approach outperforms the ST-RBM~\cite{ElaiwatBB16} and the STM-ExpLet~\cite{LiuSWC14}. On the MMI dataset, our approach outperforms the DTAGN~\cite{JungLYPK15} and the STM-ExpLet~\cite{LiuSWC14}. However, it is behind ST-RBM~\cite{ElaiwatBB16}. 

\begin{table}[ht]
\centering
\caption{Overall accuracy (\%) on CK+ and MMI datasets. Here, $^{(A)}$: appearance (or color); $^{(G)}$: geometry (or shape); $^*$: Deep Learning based approach; last row: ours}\label{tab:ARR_CK+_MMI}
  \footnotesize
\begin{tabular}{{l}|{c}|{c}}
\textbf{Method} & \textbf{CK+} & \textbf{MMI} \\
    \hline
    $^{(A)}$ 3D HOG (from \cite{JungLYPK15}) & 91.44 & 60.89\\
    $^{(A)}$ 3D SIFT  (from \cite{JungLYPK15}) & - & 64.39\\
    $^{(A)}$ Cov3D  (from \cite{JungLYPK15}) & 92.3 & - \\
    \hline
    $^{(A)}$ STM-ExpLet \cite{LiuSWC14} (10-fold)& \textbf{94.19} & \textbf{75.12}\\
    $^{(A)}$ CSPL \cite{ZhongLYLHM12} (10-fold) & 89.89 & 73.53 \\
    $^{(A)}$ F-Bases \cite{Sariyanidi2017} (LOSO) & \textbf{96.02} & \textbf{75.12}\\
    $^{(A)}$ ST-RBM \cite{ElaiwatBB16} (10-fold) & \textbf{95.66} & \textbf{81.63} \\

    $^{(A)}$ 3DCNN-DAP \cite{LiuLSWC14} $^*$ (15-fold) & 87.9 & 62.2\\
    $^{(A)}$ DTAN \cite{JungLYPK15} $^*$ (10-fold)& 91.44 & 62.45 \\
    $^{(A+G)}$ DTAGN \cite{JungLYPK15} $^*$ (10-fold)& \textbf{97.25} & \textbf{70.24} \\
    \hline
    $^{(G)}$ DTGN \cite{JungLYPK15} $^*$ (10-fold) & 92.35 & 59.02\\
    $^{(G)}$ TMS \cite{JainHA11} (4-fold) & 85.84 & - \\
    $^{(G)}$ HMM \cite{Wang2013CVPR} (15-fold) & 83.5 & 51.5 \\
    $^{(G)}$ ITBN \cite{Wang2013CVPR} (15-fold) & 86.3 & 59.7 \\
    $^{(G)}$ Velocity on $\mathcal{G}(n,2)$\cite{Taheri2001FG} & 82.8 & -\\
        \hline
    $^{(G)}$ traj. on $\mathcal{G}(2,n)$ (10-fold) & 94.25 $\pm$ 3.71 & 78.18 $\pm$ 4.87\\
    $^{(G)}$ \textbf{traj. on $\mathcal{S}^{+}(2,n)$ (10-fold)} & \textbf{96.87 $\pm$ 2.46} & \textbf{79.19 $\pm$ 4.62 }\\
    \hline
  \end{tabular}
\end{table}

\begin{table}[!htb]
\centering
\caption{Overall accuracy on Oulu-CASIA and AFEW dataset (following the EmotiW'13 protocol~\cite{DhallGJWG13})}\label{tab:ARR_AFEW}
  \footnotesize
\begin{tabular}{{l}|{c}|{c}}
\textbf{Method} & \textbf{Oulu-CASIA} & \textbf{AFEW}  \\
     \hline
$^{(A)}$ HOG 3D \cite{KlaserMS08} & 70.63 & 26.90\\
$^{(A)}$ 3D SIFT \cite{ScovannerAS07} & 55.83 & 24.87\\

$^{(A)}$ LBP-TOP \cite{ZhaoP07} & 68.13 & 25.13\\
 \hline
$^{(A)}$ EmotiW \cite{DhallGJWG13} & - & 27.27\\
$^{(A)}$ STM \cite{LiuSWC14} & - & 29.19\\
$^{(A)}$ STM-ExpLet \cite{LiuSWC14} & 74.59 & 31.73\\
$^{(A+G)}$ DTAGN \cite{JungLYPK15} $^*$ (10-fold)& \textbf{81.46} & - \\
$^{(A)}$ ST-RBM \cite{ElaiwatBB16} & - & \textbf{46.36}\\
     \hline
\textbf{$^{(G)}$ traj. on $\mathcal{G}(2,n)$} & 80.0 $\pm$ 5.22 &  39.1 \\
\textbf{$^{(G)}$ traj. on $\mathcal{S}^{+}(2,n)$} & \textbf{83.13 $\pm$ 3.86} & \textbf{39.94} \\
 \hline
  \end{tabular}
\end{table}

On the Oulu-CASIA dataset, our approach shows a clear superiority to existing methods, in particular STM-ExpLet~\cite{LiuSWC14} and DTGN~\cite{JungLYPK15}. Elaiwat~\emph{et al.}~\cite{ElaiwatBB16} do not report any results on this dataset, however, their approach achieved the highest accuracy on AFEW. Our approach is ranked second showing a superiority to remaining approaches on AFEW.

\section{Conclusion}\label{sect:conclusion}
In this paper, we have proposed a geometric approach for effectively modeling and classifying dynamic $2$D and $3$D landmark sequences for human behavior understanding. Based on Gramian matrices derived from the static landmarks, our representation consists of an affine-invariant shape representation and a spatial covariance of the landmarks. We have exploited the geometry of the space to define a closeness between static shape representations. Then, we have derived computational tools to align, re-sample and compare these trajectories giving rise to a rate-invariant analysis. Finally, landmark sequences are learned from these trajectories using a variant of SVM, called ppfSVM, which allows us to deal with the nonlinearity of the space of representation. We evaluated our approach in three different applications, namely, $3$D human action recognition, $3$D emotion recognition from body movement, and $2$D facial expression recognition. Extensive experiments on nine publicly available datasets showed that the proposed approach achieves competitive or better results than state-of-art solutions.

\section{Acknowledgements}
We thank Professor Yvonne Delevoye for fruitful discussions on emotion from 3D human dynamics. This work has been partially supported by PIA, ANR (grant ANR-11-EQPX- 0023).
\ifCLASSOPTIONcaptionsoff
  \newpage
\fi


\bibliographystyle{IEEEtran}
\nocite{*}
\bibliography{bare_jrnl_compsoc.bbl}

\newpage

\begin{center}
  \textbf{\Large Supplementary Material to the paper \\``A Novel Geometric Framework \\ on Gram Matrix Trajectories \\ for Human Behavior Understanding" \\}
  
  \vspace{8pt}
  \textbf{Anis Kacem, Mohamed~Daoudi, Boulbaba~Ben~Amor, Stefano~Berretti, and~Juan~Carlos~Alvarez-Paiva  \\}
  \vspace{8pt}

\end{center}
  \textit{In this document, we provide the mathematical preliminaries to our approach. We also give further details on the conducted experiments and the reported results.}
\setcounter{equation}{0}
\setcounter{figure}{0}
\setcounter{table}{0}
\setcounter{page}{1}
\setcounter{section}{0}
\renewcommand{\theequation}{S\arabic{equation}}
\renewcommand{\thefigure}{S\arabic{figure}}

\section{Mathematical Preliminaries}
\label{sect:math-preliminaries}
We briefly review some basics of the Grassmann manifold and the manifold of symmetric positive definite matrices.

\textbf{Grassmann manifold $\mathcal{G}(d,n)$ --}
To have a better understanding of the Grassmann manifold, we first define the Riemannian manifold of the set of $n \times d$ matrices with orthonormal columns, which is known as the Stiefel manifold $\mathcal{V}_{d,n}$. 

A Grassmann manifold $\mathcal{G}(d,n)$ is the set of the $d$-dimensional subspaces of $\mathbb{R}^{n}$, where $n>d$. A subspace $\mathcal{U}$ of $\mathcal{G}(d,n)$ is represented by an $n \times d$ matrix $U$, whose columns store an orthonormal basis of this subspace. Thus, $U$ is said to span $\mathcal{U}$, and $\mathcal{U}$ is said to be the column space (or span) of $U$, and we write $\mathcal{U} = span(U)$. 
The geometry of the Grassmannian $\mathcal{G}(d,n)$ is then easily described by the map
\begin{equation}
\lspan : \mathcal{V}_{d,n} \rightarrow \mathcal{G}(d,n) \; ,
\end{equation}

\noindent that sends an $n \times d$ matrix with orthonormal columns $U$ to their span $\lspan(U)$. 
Given two subspaces $\mathcal{U}_1 = \lspan(U_1)$ and $\mathcal{U}_2=\lspan(U_2) \in \mathcal{G}(d,n)$, the geodesic curve connecting them is
\begin{dmath}
\label{eq:GeoGrasssup}
\lspan(U(t))= \lspan(U_{1}\cos(\Theta t)+M\sin(\Theta t)) \; ,
\end{dmath}

\noindent where $\Theta=diag(\theta_1,\theta_2)$ is a $d \times d$ diagonal matrix formed by the \textit{principal angles} between $\mathcal{U}_1$ and $\mathcal{U}_2$, while the matrix $M$ is given by $M=(I_n-U_{1}U_1^T)U_2 F$, with $F$ being the pseudo-inverse $diag(\sin(\theta_1),\sin(\theta_2))$. The Riemannian distance between $\mathcal{U}_1$ and $\mathcal{U}_2$ is given by 
\begin{equation}
d^2_{\mathcal{G}}(\mathcal{U}_1,\mathcal{U}_2)=\|\Theta\|^2_F \; .
\label{eq:distGrasssup}
\end{equation}

\textbf{Manifold of Symmetric Positive Definite (SPD) matrices $\mathcal{P}_d$} -- It is known to be the positive cone in $\mathbb{R}^{d}$, and has been extensively used to study covariance matrices~\cite{tuzel:2006,sanin:2013,bhattacharya:2016}. A symmetric $d \times d$ matrix $R$ is said to be positive definite if and only if $v^{\mathrm{T}}Rv > 0$ for every non-zero vector $v \in \mathbb{R}^{d}$. $\mathcal{P}_d$ is mostly studied when endowed with a Riemannian metric, thus forming a Riemannian manifold. A number of metrics have been proposed for $\mathcal{P}_d$, the most popular ones being the Affine-Invariant Riemannian Metric (AIRM) and the log-Euclidean Riemannian  metric (LERM)~\cite{arsigny2006log}. In this study, we only consider the AIRM for its robustness~\cite{TuzelPM08}. 

With this metric, the geodesic curve connecting two SPD matrices $R_{1}$ and $R_{2}$ in $\mathcal{P}_d$ is 
\begin{equation}
R(t)=R_{1}^{1/2}\exp(t\log (R_{1}^{-1/2}R_{2}R_{1}^{-1/2}))R_{1}^{1/2} \; ,
\label{eq:spdgeo}
\end{equation}

\noindent where $log(.)$ and $exp(.)$ are the matrix logarithm and exponential, respectively. The Riemannian distance between $R_{1}$ and $R_{2}$ is given by

\begin{equation}
d^2_{P_d}(R_1,R_2)=\| \log{(R_1^{-1/2}R_2R_1^{-1/2})}\|^2_{\textrm{F}} \; ,
\label{eq:spddist}
\end{equation}

\noindent where $\lVert . \rVert_{F}$ denotes the Frobenius matrix norm.

For more details about the geometry of the Grassmannian $\mathcal{G}(d,n)$ and the positive definite cone ${\mathcal{P}_d}$, readers are referred to~\cite{absil2004riemannian,BegelforW06,Bonnabel2009SIAM,Pennec2006IJCV}.

\section{Body Part Decomposition}
\begin{figure}[ht]
\centering
\includegraphics[width=70mm,scale=0.7]{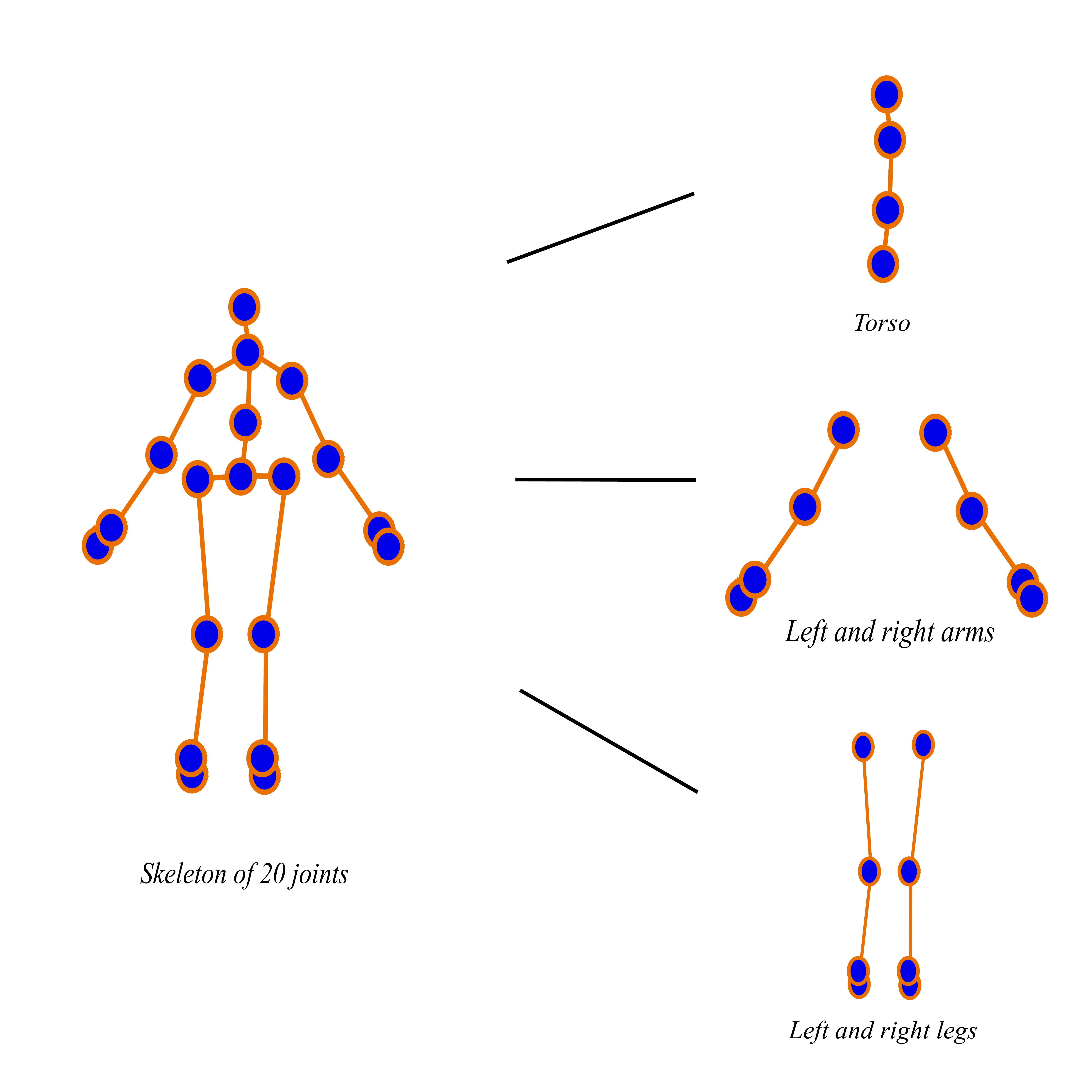}
\centering
\caption{Decomposition of the Kinect skeleton into three body parts.}
\label{Fig:bodyparts}
\end{figure}

In Fig.~\ref{Fig:bodyparts}, we show an example of the proposed Kinect skeleton decomposition into three body parts (Section~6.1). 

\begin{figure}[ht]
\centering
\includegraphics[width=70mm,scale=0.7]{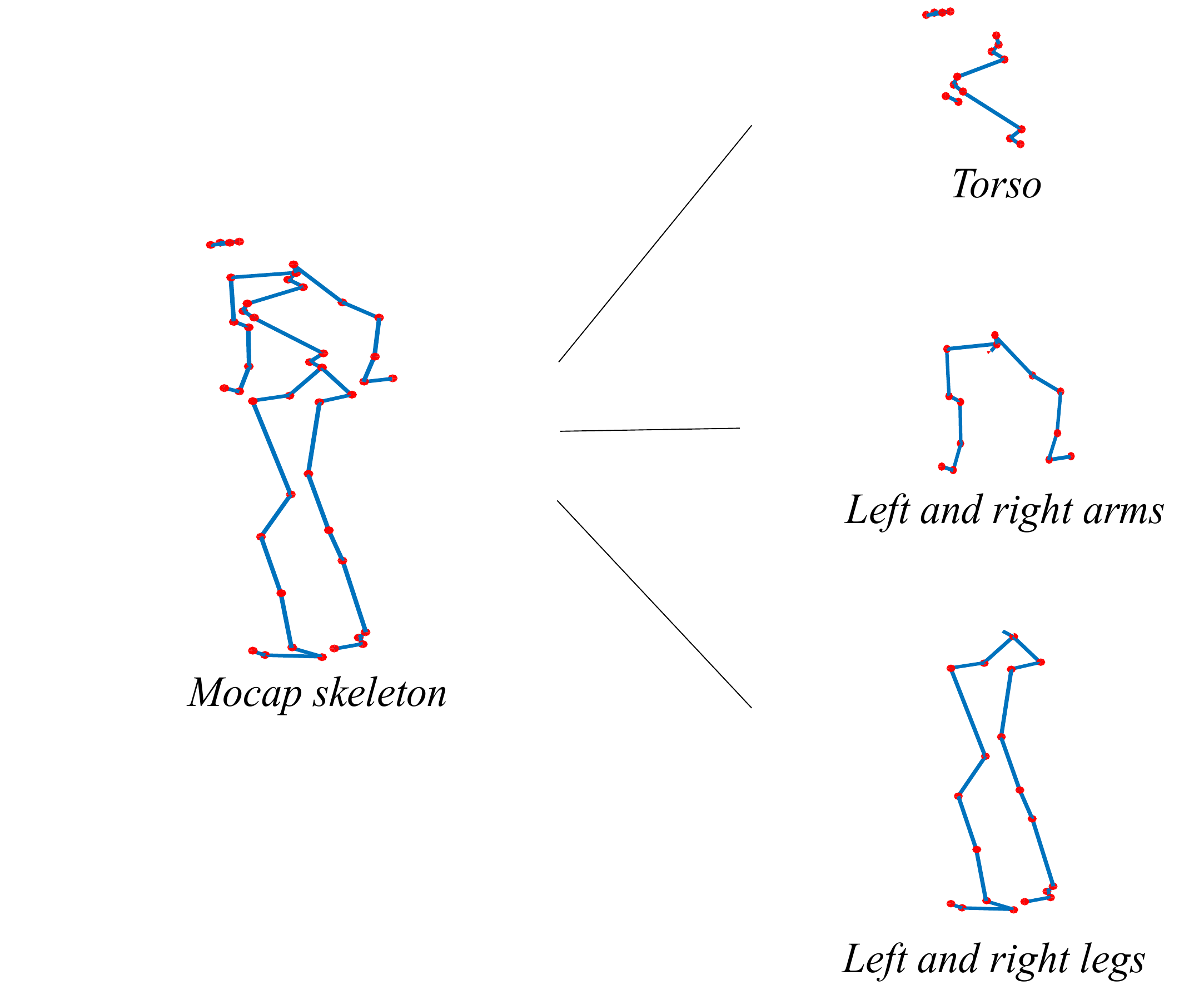}
\centering
\caption{Decomposition of the MoCap skeleton into three body parts.}
\label{Fig:bodypartsmocap}
\end{figure}

An example of the same decomposition on MoCap skeletons for emotion recognition from body movement (Section~6.2) is shown in Fig.~\ref{Fig:bodypartsmocap}.
 
\section{Action Recognition Results}
As supplement to the results of our approach in the $3$D action recognition task (Section~6.1.3), we report the obtained confusion matrices on the four datasets used in the experiments.

\begin{figure}[ht]
\centering
\includegraphics[width=.86\linewidth]{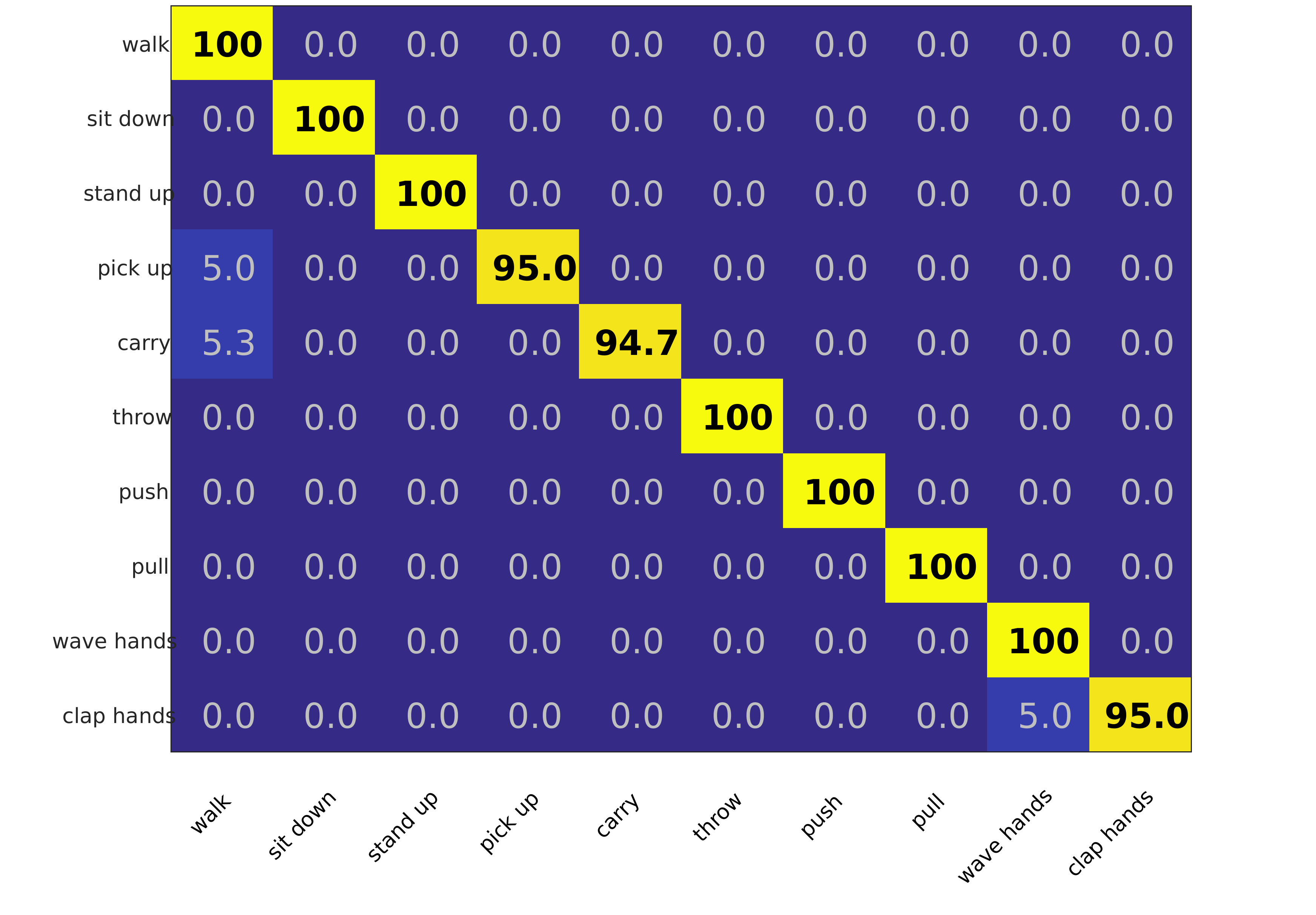}
\centering
\caption{Confusion matrix for the UT-Kinect dataset.}
\label{Fig:confusion_utkinect}
\end{figure}

In Fig.~\ref{Fig:confusion_utkinect}, we show the confusion matrix for the UT-Kinect dataset. We can observe that all the actions were well recognized. The few confusions happened between \textit{``pick up''} with \textit{``walk''}, \textit{``carry''} with \textit{``walk''}, and \textit{``clap hands''} with \textit{``wave hands''}.

\begin{figure}[ht]
\centering
\includegraphics[width=.86\linewidth]{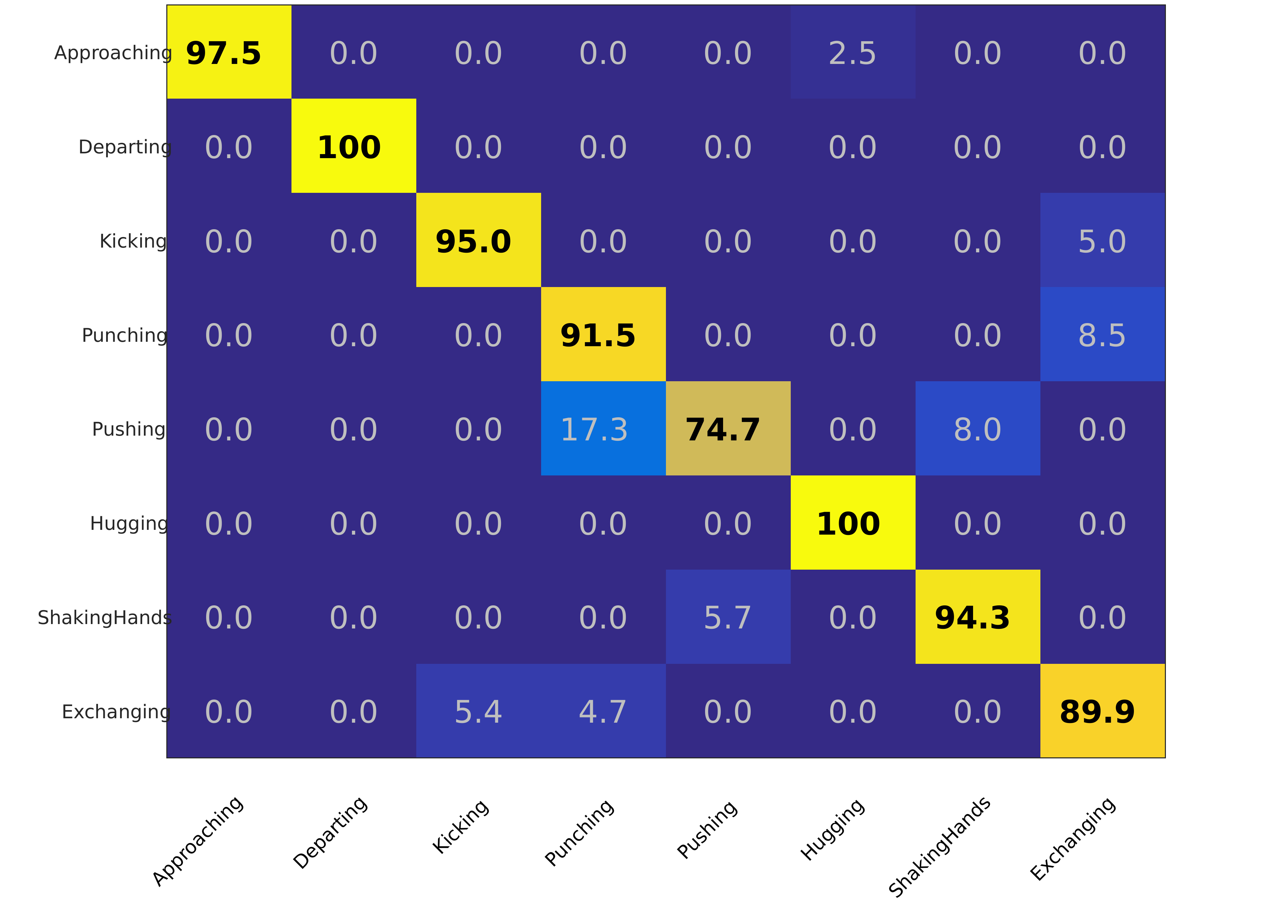}
\centering
\caption{Confusion matrix for the SBU dataset.} 
\label{Fig:confusion_sbu}
\end{figure}

On the human interaction SBU dataset, as shown in Fig.~\ref{Fig:confusion_sbu}, the highest performance was achieved for \textit{``departing''} and \textit{``hugging''} interactions ($100\%$), while \textit{``pushing''} interaction was the least recognized ($74.7\%$). The latter was mainly confused by our approach with a similar interaction (\emph{i.e.}, \textit{``punching''}).

\begin{figure}[ht]
\centering
\includegraphics[width=.95\linewidth]{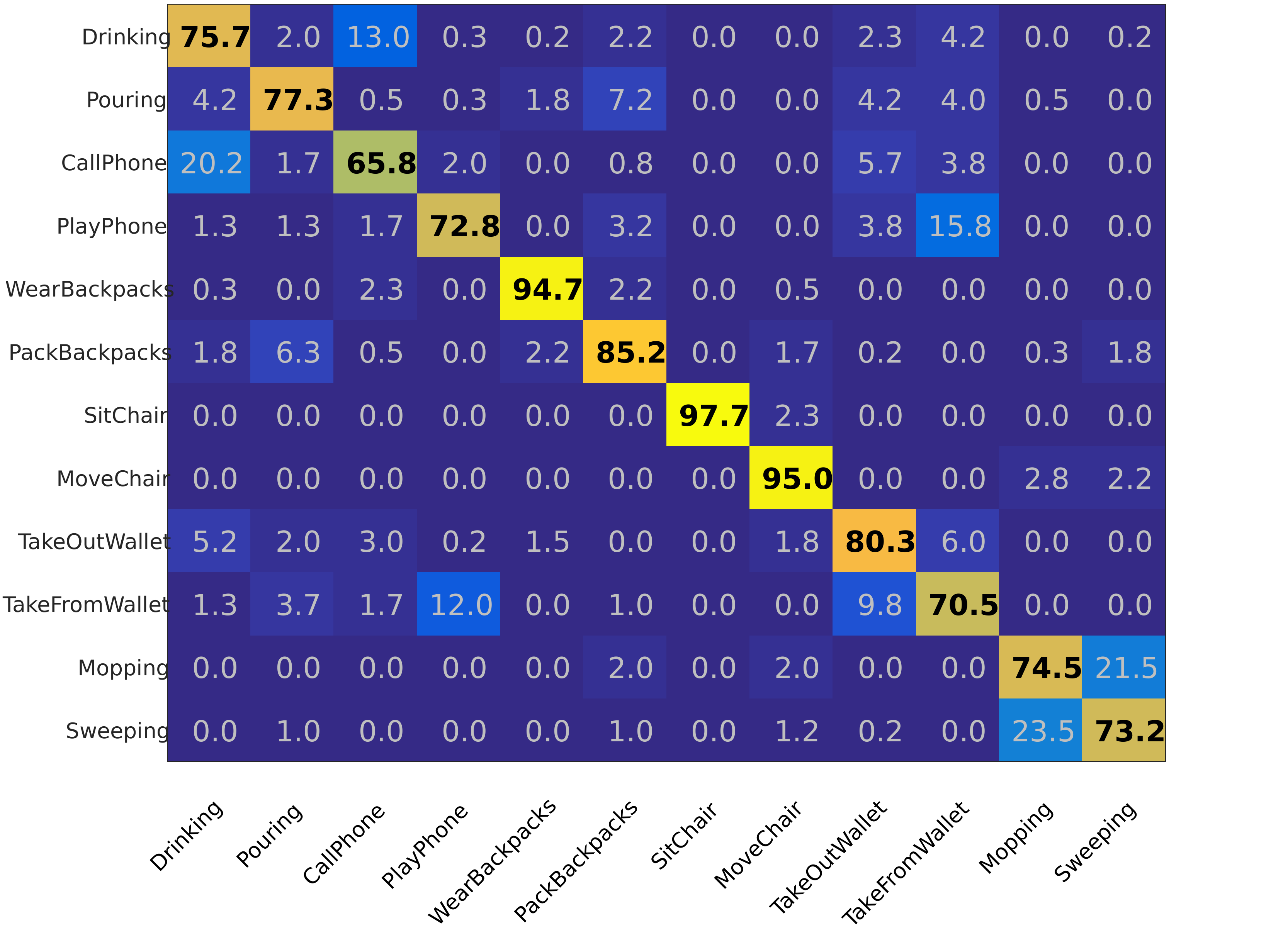}
\centering
\caption{Confusion matrix for the SYSU3D dataset.} 
\label{Fig:confusion_sysu}
\end{figure}

Figure~\ref{Fig:confusion_sysu} depicts the confusions of our approach on the human-object interaction dataset SYSU$3$D. Unsurprisingly, \textit{``sit chair''} and \textit{``move chair''} were the most recognized interactions ($>95\%$). In accordance with~\cite{hu2015jointly}, the lowest performance was achieved for \textit{``call phone''} interaction ($65.8\%$), which was mutually confused with \textit{``drinking''}. These two interactions involve similar patterns (raising one arm to the head) that could be more similar with the inaccurate tracking of the skeletons. Other examples of such mutual confusions include the interactions \textit{``take from wallet''} ($70.5\%$) with \textit{``play phone''} ($72.8\%$) and \textit{``mopping''} ($74.5\%$) with \textit{``sweeping''} ($73.2\%$).

\begin{figure}[ht]
\centering
\includegraphics[width=.95\linewidth]{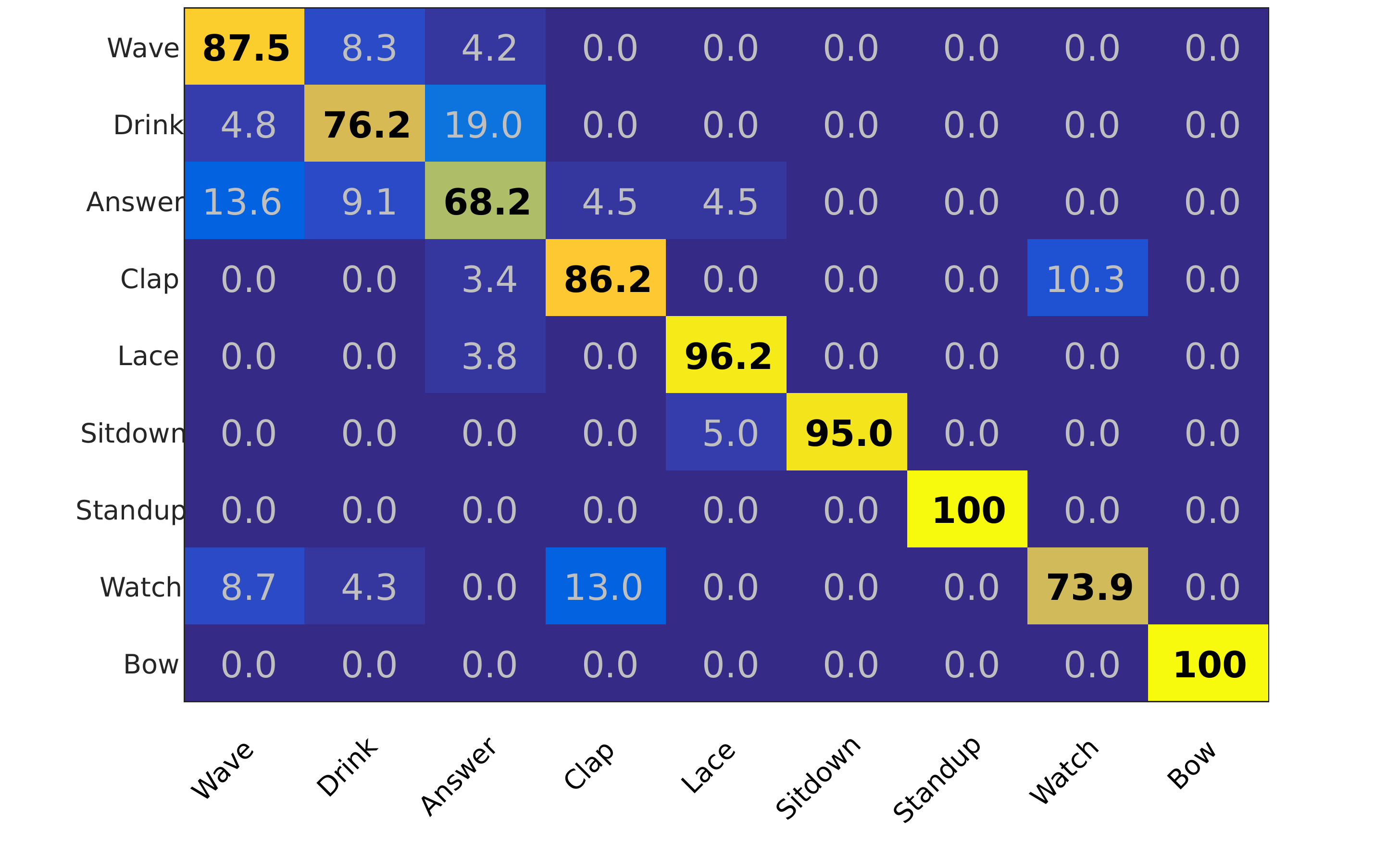}
\centering
\caption{Confusion matrix for the Florence dataset.} 
\label{Fig:confusion_florence}
\end{figure}

Finally, we report in Fig.~\ref{Fig:confusion_florence} the confusion matrix for the Florence $3$D dataset. Similarly to the reported results on the UT-Kinect dataset, the best performance was recorded for the \textit{``stand up''} ($100\%$) and \textit{``sit down''} ($95\%$) actions. Correspondingly to the obtained results on the SYSU3D dataset, the main confusions concerned \textit{``drink''} ($76.2\%$) with \textit{``answer phone''} ($68.2\%$). Furthermore, it is worth noting that, in this dataset, several actions are performed with the right arm by some participants, while others acted it with the left arm. This could explain the low performance achieved by our approach on distinguishing \textit{``read watch''}, where only one arm (left or right) is raised to the chest, from \textit{``clap hands''}, where the two arms are raised to merely the same position.

\begin{figure*}[ht]
\centering
\includegraphics[width=.85\linewidth]{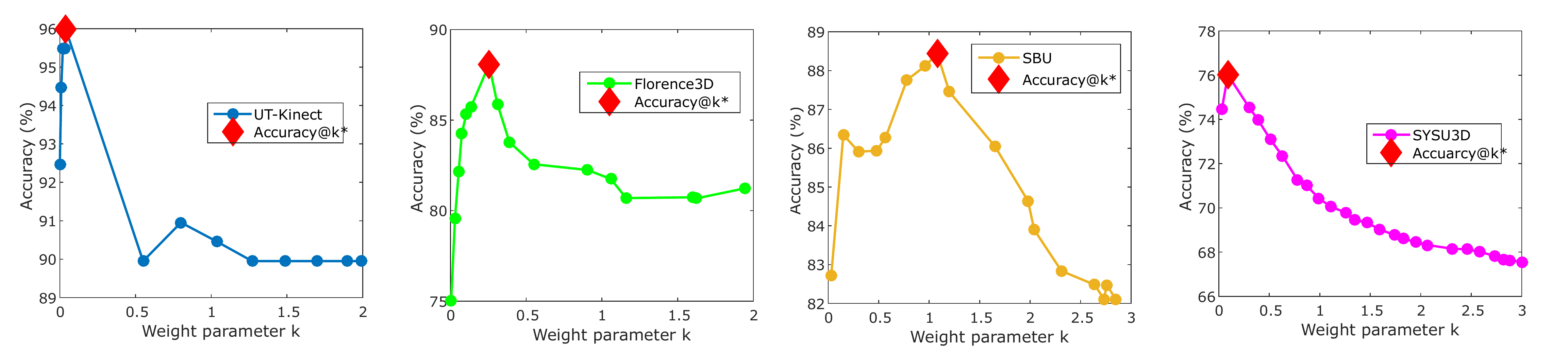}
\caption{Accuracy of the proposed approach when varying the weight parameter $k$: results for the UT-Kinect, Florence3D, SBU, and SYSU-3D datasets are reported from left to right.}
\label{Fig:Action_Acc_vs_k}
\end{figure*}

Our approach involves an important parameter that controls the contribution of the spatial covariance with respect to the affine-invariant information on the Grassmannian. In Fig.~\ref{Fig:Action_Acc_vs_k}, we report the accuracies obtained when considering the whole skeletons for different values of $k$. The optimal values are $k^*= 0.05$, $k^*=0.81$, $k^*=0.25$, and $k^*=0.09$ for the UT-Kinect, SBU, Florence3D, and SYSU3D datasets, respectively.

\section{Emotion Recognition from Body Movement Results}
\begin{figure}[t!]
\centering
\includegraphics[width=\linewidth]{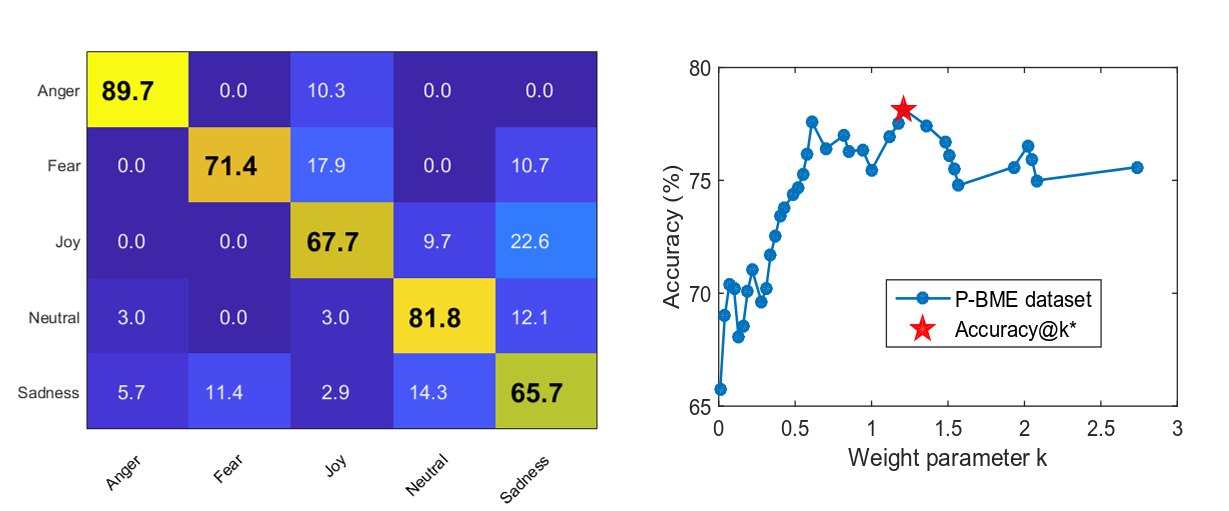}
\caption{\label{fig:confimpactk} P-BME dataset: Confusion matrix (left). Impact of the parameter $k$ on emotion recognition accuracy (right).}
\end{figure}

In Fig.~\ref{fig:confimpactk} (left), we report the confusion matrix of different emotions. The diagonal dominance of the matrix can be observed with the best results scored by \textit{neutral} and \textit{anger} (more than $80\%$), followed by \textit{fear} ($71\%$), \textit{joy} (about $67\%$), with the lowest accuracy for \textit{sadness} (about 65\%).
In Fig.~\ref{fig:confimpactk} (right), we report the obtained results for $k \in [0,3]$ with a step of $0.1$.

In Table~\ref{tab:parts-classification}, we report the obtained accuracies per emotion for each body part. With this evaluation, we are able to identify body parts that are more informative to a specific emotional state. We can observe that \textit{Anger}, \textit{Fear}, and \textit{Joy} are better recognized with the whole body, while \textit{Neutral} and \textit{Sadness} are better recognized with arms. 
One can note that the performance for these two emotions increases after body part fusion compared to the whole body only, notably through the contribution of arms.

\begin{table}[ht]
\centering
\caption{\label{tab:parts-classification} Comparative study of emotion recognition (\%) on the P-BME dataset using different parts of the body and our proposed method. Anger (An), Fear (Fe), Joy (Jo), Neutral (Ne), Sadness (Sa), Accuracy (Acc)}
  \footnotesize
\begin{tabular}{{l}|{c}|{c}|{c}|{c}|{c}||{c}}
\textbf{Method} & \textbf{An} & \textbf{Fe} & \textbf{Jo} & \textbf{Ne}  & \textbf{Sa} & \textbf{Acc} \\
   \hline

Legs only & 55.1 & \textbf{64.3} & 35.5 & 57.6 & 60 & 59.17 \\
Arms only & 55.2 & 57.1 & 45.2 & \textbf{84.8} & \textbf{71.4} & 69.42  \\ 
Torso only & \textbf{82.76} & 50 & 48.4 & \textbf{75.7} & 54.3 &  67.23 \\
Full body & \textbf{89.6} & \textbf{78.5} & \textbf{58.0} & \textbf{72.7} & \textbf{65.7} & \textbf{78.15} \\
Late BP Fusion & \textbf{89.7} & \textbf{71.4} & \textbf{67.7} & \textbf{81.8} & \textbf{65.7}  & \textbf{81.99} \\
 \hline
  \end{tabular}
\end{table}

Finally, we evaluated our approach when considering subsequences of the original sequence. In Table~\ref{tab:aclengthseq}, we provide the obtained results and the execution time of the testing phase, when considering only $25\%$, $50\%$, $75\%$, and $100\%$ of the sequence. The execution time is recorded for a test sequence of $1,118$ frames (about $8$ seconds) when considering separately the four temporal subsequences. The highest execution time is about $2$ seconds, which is satisfactory considering the high frame-rate of the data. Unsurprisingly, the best accuracy is obtained when considering the whole sequence. The performance decreases when shorter subsequences are used to perform emotion recognition.  

\begin{table}[ht]
\centering
\caption{Emotion recognition accuracy using different sequence lengths on the P-BME dataset}\label{tab:aclengthseq}
\footnotesize
\begin{tabular}{{l}|{c}|{c}}
\textbf{Sequence length} & \textbf{Accuracy (\%)} & \textbf{Exec. time (s)}\\
    \hline
    25\% of the sequence & 61.20 $\pm$ 7.52 &    1.90 \\
    50\% of the sequence &  67.27 $\pm$ 6.36 &   1.93 \\
    75\% of the sequence & 70.88 $\pm$ 6.81 &    1.95 \\
    \textbf{100\% of the sequence} & \textbf{78.15 $\pm$ 5.79 } &    \textbf{1.99} \\
     \hline
  \end{tabular}
\end{table}

\section{Facial Expression Recognition Results}
Here, we provide further details on the obtained results in the task of $2$D facial expression recognition. Specifically, we supply the confusion matrices for the considered datasets. We also report the obtained results with respect to baseline experiments that are quite similar to those considered in $3$D action recognition (see Section~6.1.3).

\begin{figure}[ht]
\centering
\includegraphics[width=\linewidth]{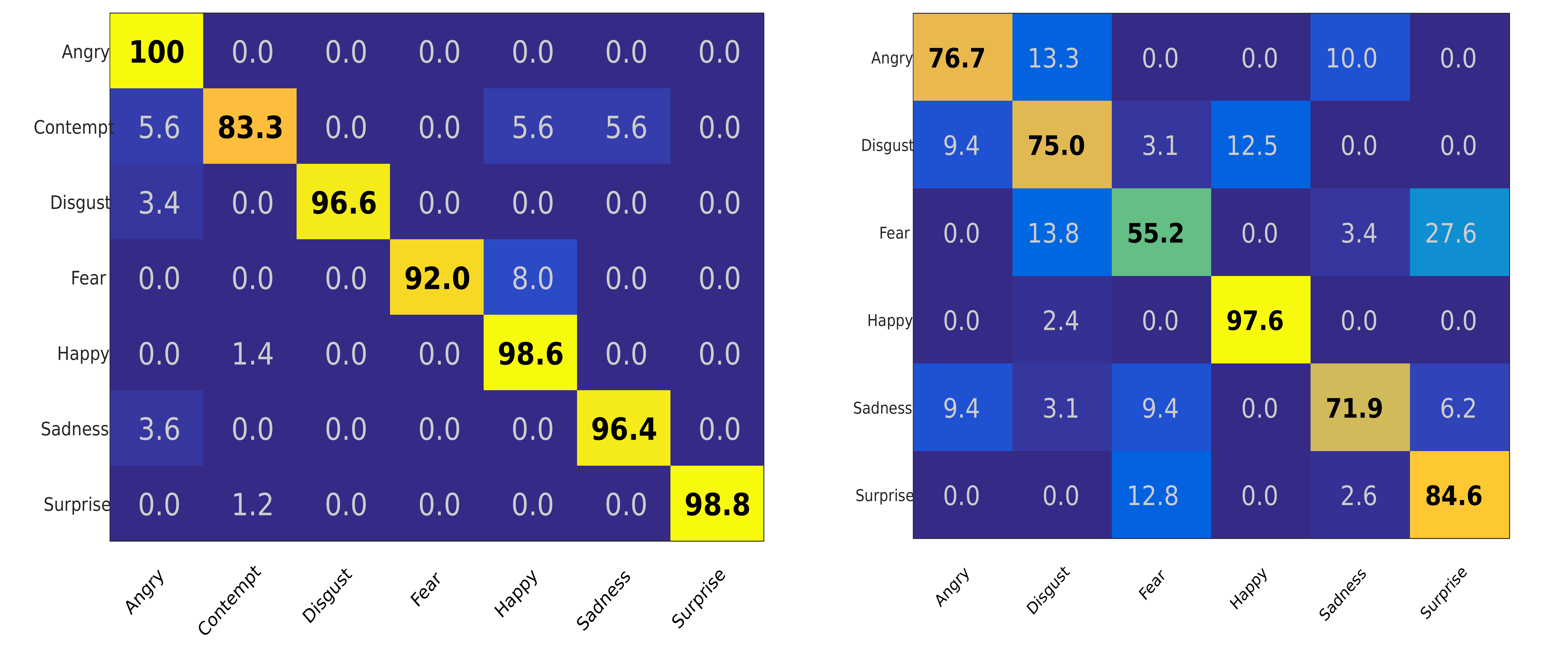}
\centering
\caption{Confusion matrices on the CK+ (left) and MMI (right) datasets.}
\label{Fig:confusion_fer1}
\end{figure}

In the left panel of Fig.~\ref{Fig:confusion_fer1}, we show the confusion matrix on the CK+ dataset. While individual accuracies of \textit{``anger''}, \textit{``disgust''}, \textit{``happiness''}, and \textit{``surprise''} are high (more than $96\%$), recognizing \textit{``contempt''} and \textit{``fear''} is still challenging (less than $92\%$). In the right panel of the same figure, we can observe that the best accuracy on the MMI dataset was also achieved for \textit{``happiness''} followed by \textit{``surprise''}. Also in this case, the lowest performance was recorded for \textit{``fear''} expression.

\begin{figure*}[ht]
\centering
\includegraphics[width=.85\linewidth]{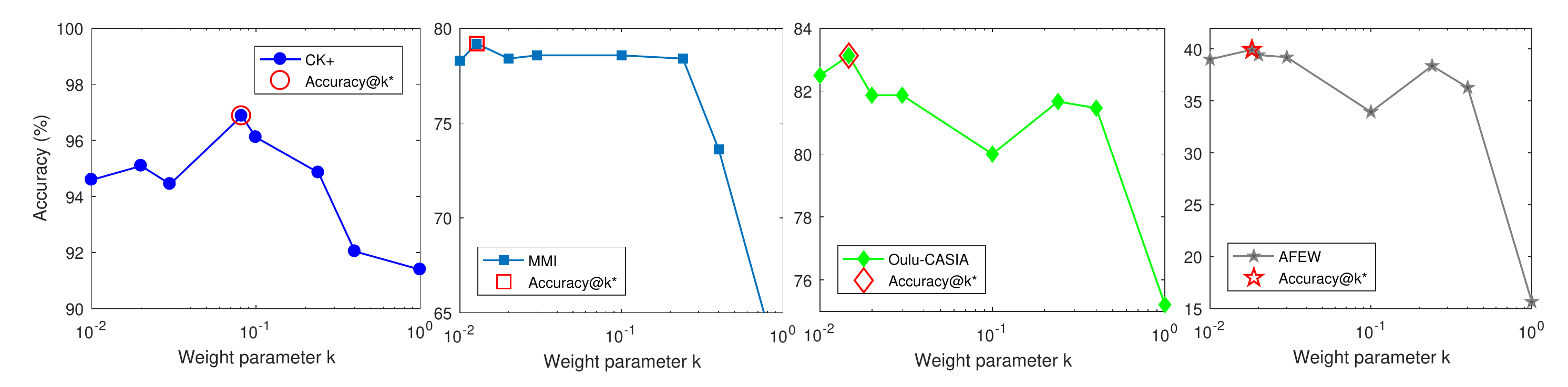}
\caption{Accuracy of the proposed approach when varying the weight parameter $k$ on, from left to right, CK+, MMI, Oulu-CASIA and AFEW.}
\label{Fig:Acc_vs_k}
\end{figure*}

\begin{figure}[ht]
\centering
\includegraphics[width=.95\linewidth]{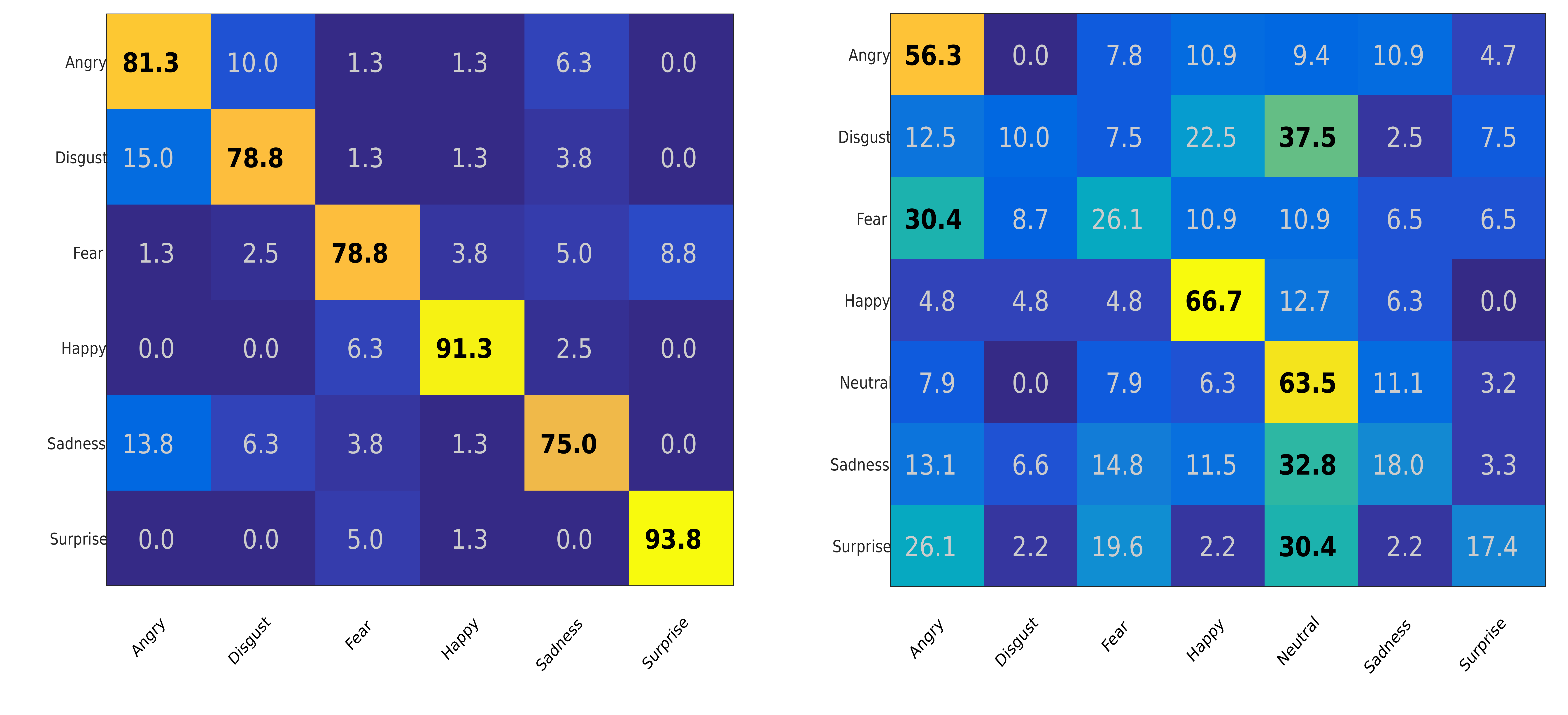}
\centering
\caption{Confusion matrices on the Oulu-CASIA (left) and AFEW (right) datasets.}
\label{Fig:confusion_fer2}
\end{figure}

As shown in Fig.~\ref{Fig:confusion_fer1}, on the Oulu-CASIA dataset the highest performance was reached for \textit{``happiness''} ($91.3\%$) and \textit{``surprise''} ($93.8\%$) expressions; \textit{``Disgust''}, \textit{``fear''}, and \textit{``sadness''} were the most challenging expressions in this dataset ($<79\%$). 
Unsurprisingly for the AFEW dataset, the \textit{``neutral''} ($63.5\%$), \textit{``anger''} ($56.3\%$), and \textit{``happiness''} ($66.7\%$) expressions are better recognized over the rest (see the right confusion matrix in Fig.~\ref{Fig:confusion_fer2}).

It is important to note that the \textit{``fear''} expression was the most challenging expression in all the datasets. In fact, this expression involves several action unit activations (\emph{i.e.}, AU1+AU2+AU4+AU5+AU7+AU20+AU26)~\cite{friesen1983emfacs} that are quite difficult to detect by using only geometric features.

\textbf{Baseline Experiments.} 
Based on the results reported in Table~\ref{tab:baseline}, we discuss in this paragraph algorithms and their computational complexity with respect to baselines. 
Firstly, we studied the computational cost of the proposed framework in the task of 2D facial expression recognition on the CK+ dataset. Correspondingly to 3D action recognition settings, we report in the top of Table~\ref{tab:baseline} the running time statistics for trajectory construction, comparison of trajectories, and the testing phase of trajectory classification in $\mathcal{S}^+(2,n)$.

\begin{table}[ht]
\centering
\caption{Baseline experiments and computational complexity on the CK+, MMI and AFEW datasets}\label{tab:baseline}
\footnotesize
\begin{tabular}{{l}|{c}}
\textbf{Pipeline component} & \textbf{Time} (s) \\
\hline 
Trajectory construction in $\mathcal{S}^+(2,n)$& 0.007 \\
Comparison of trajectories in $\mathcal{S}^+(2,n)$& 0.055 \\
Classification of a trajectory in $\mathcal{S}^+(2,n)$& 6.28 \\
\hline
\end{tabular}
  
\vspace{0.3cm}

\begin{tabular}{{l}|{c}|{c}}
\textbf{Distance} & \textbf{CK+} (\%) & \textbf{Time} (s) \\
\hline 
Flat distance $d_{\mathcal{F}^+}$ & 93.78 $\pm$ 2.92 & 0.020 \\
Distance $d_{\mathcal{P}_n}$ in $\mathcal{P}_n$  & 92.92 $\pm$ 2.45 & 0.816 \\
Closeness $d_{\mathcal{S}^+}$ & \textbf{96.87$\pm$ 2.46}  & 0.055 \\
\hline
\end{tabular}
  
\vspace{0.3cm}  
\begin{tabular}{{l}|{c}|{c}|{c}}
\textbf{Temporal alignment} & \textbf{CK+ (\%)} & \textbf{MMI (\%)} & \textbf{Time} (s) \\
\hline
without DTW & 90.94 $\pm$ 4.23 & 66.93 $\pm$ 5.79 & 0.018 \\
with DTW &  \textbf{96.87 $\pm$ 2.46} & \textbf{79.19 $\pm$ 4.62} & 0.055\\
\hline
\end{tabular}

\vspace{0.3cm}  
\begin{tabular}{{l}|{c}|{c}}
\textbf{Adaptive re-sampling} & \textbf{MMI (\%)} & \textbf{AFEW (\%)} \\
\hline
without resampling & 74.72 $\pm$ 5.34 & 36.81  \\
with resampling &  \textbf{79.19 $\pm$ 4.62} & \textbf{39.94}\\
\hline
\end{tabular}
\vspace{0.3cm}      
  
\begin{tabular}{{c}|{c}|{c}}
\textbf{Classifier} & \textbf{CK+ (\%)} & \textbf{AFEW (\%)} \\
\hline 
K-NN & 88.97 $\pm$ 6.14 & 29.77 \\
ppf-SVM & \textbf{96.87 $\pm$ 2.46} & \textbf{39.94} \\
\hline
\end{tabular}
\end{table}

Then, we have used different distances defined on $\mathcal{S}^+(2,n)$. Specifically, given two matrices $G_1$ and $G_2$ in $\mathcal{S}^+(2,n)$: (1) we used $d_{\mathcal{P}_n}$ to compare them by regularizing their ranks, \emph{i.e.}, making them $n$ full-rank, and considering them in $\mathcal{P}_n$ (the space of $n$-by-$n$ positive definite matrices), $d_{\mathcal{P}_n}(G_1,G_2)=d_{\mathcal{P}_n}(G_1+\epsilon I_n,G_2+\epsilon I_n)$; (2) we used the Euclidean flat distance $d_{\mathcal{F}^+}(G_1,G_2)=\|G_1-G_2\|_F$, where $\|.\|_F$ denotes the Frobenius-norm. The closeness $d_{\mathcal{S}^+}$ between two elements of $\mathcal{S}^+(2,n)$ defined in Eq.~(7) is more suitable, compared to the distance $d_{\mathcal{P}_n}$ and the flat distance $d_{\mathcal{F}^+}$ defined in literature. This demonstrates the importance of being faithful to the geometry of the manifold of interest. Another advantage of using $d_{\mathcal{S}^+}$ over $d_{\mathcal{P}_n}$ is the computational time, as it involves $n$-by-$2$ and $2$-by-$2$ matrices instead of $n$-by-$n$ matrices.  
Note that the provided execution times are relative to the comparison of two arbitrary sequences. 

Table~\ref{tab:baseline} reports the average accuracy when DTW in used or not in our pipeline, on both the CK+ and MMI datasets. It is clear from these experiments that a temporal alignment of the trajectories is a crucial step, as an improvement of about $12\%$ is obtained on MMI and of approximately $6\%$ on CK+. 

The adaptive re-sampling tool is also analyzed. When it is included in the pipeline, an improvement of about $5\%$ is achieved on MMI and $3\%$ on AFEW. 

In the last Table, we compare the results of ppfSVM with a $k$-Nearest Neighbor classifier for both the CK+ and AFEW datasets. The number of nearest neighbors $k$ to consider for each dataset is chosen by cross-validation. On CK+, we obtained an average accuracy of $88.97\%$ for $k=11$. On AFEW, we obtained an average accuracy of $29.77\%$ for $k=7$. These results are outperformed by the ppfSVM classifier.

Finally, in Fig.~\ref{Fig:Acc_vs_k}, we study the method when varying the parameter $k$ (closeness) defined in Eq.~(7). The graphs report the method accuracy on CK+, MMI, Oulu-CASIA, and AFEW, respectively. 

  \newpage

\vskip -25pt plus -1fil
\begin{IEEEbiography}
[{\includegraphics[width=1in,height=1.25in,clip,keepaspectratio]{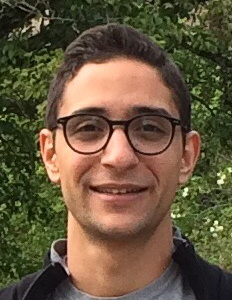}}]{Anis Kacem} received
his engineering degree in applied sciences and technology in 2014, from the National Institute of Applied Sciences and Technology (INSAT), Tunisia. Currently, he is a Ph.D. candidate in Computer Science at Institut Mines T\'el\'ecom (IMT) Lille-Douai, France, working with the 3DSAM research team of CRIStAL laboratory (CNRS UMR 9189). His research interests are mainly focused on pattern recognition and shape analysis with applications to facial expression recognition and human behavior understanding.
\end{IEEEbiography}
\vskip -25pt plus -1fil
\begin{IEEEbiography}[{\includegraphics[width=1in,height=1.25in,clip,keepaspectratio]{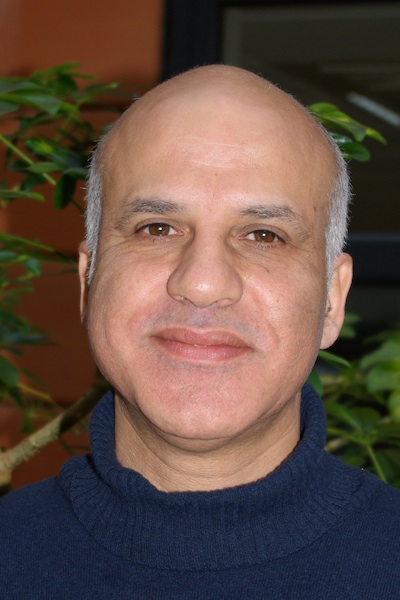}}]{Mohamed Daoudi}
is a Full Professor of Computer Science at IMT Lille Douai and the Head of Image group at CRIStAL Laboratory (UMR CNRS 9189). He received his Ph.D. degree in Computer Engineering from the University of Lille 1 (France) in 1993 and Habilitation Ã Diriger des Recherches from the University of Littoral (France) in 2000. His research interests include pattern recognition, shape analysis and computer vision. He has published over 150 papers in some of the most distinguished scientific journals and international conferences. He is Associate Editor of Image and Vision Computing Journal and IEEE Transactions on Multimedia. He is a General co-Chair of IEEE International Conference on Automatic Face and Gesture Recognition 2019. He is Fellow of IAPR and IEEE Senior member.
\end{IEEEbiography}
\vskip -25pt plus -1fil
\begin{IEEEbiography}[{\includegraphics[width=1in,height=1.25in,clip,keepaspectratio]{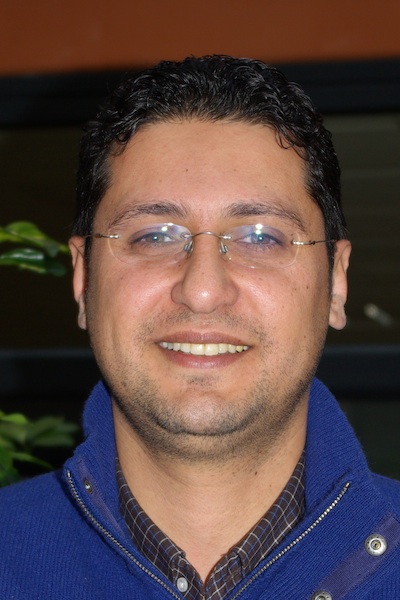}}]{Boulbaba Ben Amor} is full professor of Computer Science with the Mines-Telecom Institute (IMT) Lille Douai and member of the UMR CNRS Research Center CRIStAL, in France. He is recipient of the prestigious Fulbright scholarship (2016-2017) and he is senior member of the IEEE, since 2015. He holds an Habilitation from University of Lille (France). He earned his Ph.D. in Computer Science from the Ecole Centrale de Lyon, in 2006. His research topics lie to 3D shape and their dynamics analysis for advanced human behavior analysis and assessment.
\end{IEEEbiography}
\vskip -25pt plus -1fil
\begin{IEEEbiography}
[{\includegraphics[width=1in,height=1.25in,clip,keepaspectratio]{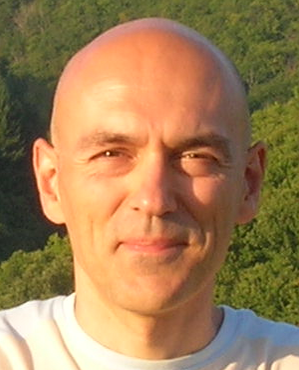}}]{Stefano Berretti} is an Associate Professor at the University of Florence. He received the Ph.D. in Computer Engineering in 2001. His research interests focus on image content modeling, indexing and retrieval, 3D computer vision for face biometrics, human emotion and behavior understanding.
On these themes, he has published over 140 articles in some of the most distinguished conferences and journals in the area of pattern recognition, computer vision and multimedia. He is the Information Director of the ACM Transactions on Multimedia Computing, Communications, and Applications, and Associate Editor of the IET Computer Vision journal.
\end{IEEEbiography}
\vskip -25pt plus -1fil
\begin{IEEEbiography}[{\includegraphics[width=1in,height=1.25in,clip,keepaspectratio]{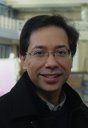}}]{Juan Carols Alvarez-Paiva}
is a Full Professor of Mathematics at University of Lille, France, and member of \textit{Laboratoire Painlev\'e} (Lille 1/CNRS UMR 8524), since 2007. He  received his Ph.D. degree in Mathematics in 1995 from Rutgers University, New Brunswick. His research interests lie in the area of differential geometry.
\end{IEEEbiography}

\end{document}